\documentclass{article} % For LaTeX2e
\usepackage{iclr2026_conference,times}

\usepackage{amsmath,amsfonts,bm}

\def\eqref#1{equation~\ref{#1}}
\def\1{\bm{1}}

\DeclareMathAlphabet{\mathsfit}{\encodingdefault}{\sfdefault}{m}{sl}
\SetMathAlphabet{\mathsfit}{bold}{\encodingdefault}{\sfdefault}{bx}{n}

\usepackage[utf8]{inputenc} % allow utf-8 input
\usepackage[T1]{fontenc}    % use 8-bit T1 fonts
\usepackage{hyperref}       % hyperlinks
\usepackage{url}            % simple URL typesetting
\usepackage{booktabs}       % professional-quality tables
\usepackage{amsfonts}       % blackboard math symbols
\usepackage{nicefrac}       % compact symbols for 1/2, etc.
\usepackage{microtype}      % microtypography
\usepackage{xcolor}         % colors
\usepackage{amsmath}
\usepackage{booktabs}
\usepackage{graphicx}
\usepackage{xspace}
\usepackage{colortbl}
\usepackage{wrapfig}
\usepackage{multirow}
\usepackage{makecell}
\usepackage{enumerate}
\usepackage{listings}
\usepackage[vlined, ruled]{algorithm2e}
\usepackage{algorithmic}
\usepackage{amsfonts,amssymb}
\usepackage{mathrsfs}
\usepackage{subcaption}
\usepackage{natbib}
\setcitestyle{numbers,square}
\usepackage{cleveref}
\usepackage{paralist, tabularx}
\usepackage{inconsolata}
\usepackage{pgfplots}
\usepackage{booktabs}
\usepackage{epigraph}
\usepackage{float}
\usepackage[normalem]{ulem}
\useunder{\uline}{\ul}{}
\usepackage[framemethod=TikZ]{mdframed}
\usepackage{framed}
\usepackage{mathtools}
\usepackage{array}
\usepackage{amsthm}
\usepackage{verbatim} 
\usepackage{bbm}
\usepackage{commath}
\usepackage{amsbsy}
\usepackage[shortlabels]{enumitem}
\usepackage{tcolorbox}
\usepackage{bbding}
\usepackage{threeparttable}
\usepackage{lipsum}
\usepackage{minitoc}
\usepackage{tocloft}

\newif\ifshowcomment
\showcommentfalse % To hide comments

\newcommand{\model}{\texttt{TDRM}\xspace}
\newcommand{\vpara}[1]{\vspace{0.05in}\noindent \textbf{#1 }}

\definecolor{custom_blue}{RGB}{49,117,181}
\definecolor{table-blue}{RGB}{173, 216, 230}
\definecolor{darkred}{RGB}{176, 36, 24}
\definecolor{cyan}{RGB}{0, 139, 139}

\title{\model: Smooth Reward Models with Temporal Difference for LLM RL and Inference}

\author{Dan Zhang$^{1\ast\dagger}$, Min Cai$^{2\ast\dagger}$, Jonathan Light$^3$, Ziniu Hu$^3$, Yisong Yue$^3$, Jie Tang$^{1,4}$\\
    $^1$\textmd{Department of Computer Science and Technology, Tsinghua University}; \\ $^2$University of Alberta; $^3$California Institute of Technology; \\ $^4$School of Electronics and Computer Science, University of Southampton
}

\iclrfinalcopy % Uncomment for camera-ready version, but NOT for submission.
\begin{document}

\maketitle

\renewcommand{\thefootnote}{\fnsymbol{footnote}}
    \footnotetext[1]{Work done when interned at Z.ai.}
    \footnotetext[2]{Equal contributions.}
    
\begin{abstract}
Reward models are central to both reinforcement learning (RL) with language models and inference-time verification. 
However, existing reward models often lack temporal consistency, leading to ineffective policy updates and unstable RL training.
We introduce \model, a method for learning smoother and more reliable reward models by minimizing temporal differences (TD) for training-time reinforcement learning and inference-time verification. 
Experiments show that TD-trained process reward models (PRMs) improve performance across Best-of-$N$ (up to 6.6\%) and tree-search (up to 23.7\%) settings.
When combined with Reinforcement Learning with Verifiable Rewards (RLVR), TD-trained PRMs lead to more data-efficient RL --- achieving comparable performance with just 2.5k data to what baseline methods require 50.1k data to attain --- and yield higher-quality language model policies in 8 model variants (5 series), e.g., Qwen2.5-(0.5B, 1,5B), GLM4-9B-0414, GLM-Z1-9B-0414, Qwen2.5-Math-(1.5B, 7B), and DeepSeek-R1-Distill-Qwen-(1.5B, 7B).
% All code is available at \url{https://anonymous.4open.science/r/TDRM-CDD6}.
We release all code at \url{https://github.com/THUDM/TDRM}.
\end{abstract}

\section{Introduction}
Reward Models (RMs), which provide rewards for the intermediate/final reasoning processes of Large Language Models (LLMs)~\cite{achiam2023gpt, team2023gemini, glm2024chatglm}, have now become a standard practice for LLM reasoning in post-training~\cite{ouyang2022training, wang2023math, zhang2025parameter}, demonstrating remarkable performance in various fields, including mathematical problem-solving~\cite{zhang2024rest, sciglm}, code synthesis~\cite{codegeex, xia2024scenegenagent}, and instruction following~\cite{cheng2024spar, liu2025dsgrm}. 
In particular, in mathematical reasoning, extensive research has explored how RMs benefit from fine-grained supervision at intermediate reasoning steps, giving rise to Process Reward Models (PRMs)~\cite{lightman2023let} that leverage such step-wise signals, as opposed to Outcome Reward Models (ORMs)~\cite{lightman2023let} relying solely on final-answer correctness.
Reward models offer important advantages: \textit{(1) RMs provide low-cost feedback signals compared to expensive human annotations, (2) PRMs enable intermediate-stage reward beyond the typically sparse signals from human or rule-based verifiers}.
Furthermore, during online Reinforcement Learning (RL) training \cite{shao2024deepseekmath}, process or rule-based reward mechanisms are crucial in enhancing LLM performance by providing effective feedback that guides reasoning quality.

However, a key limitation of current RMs lies in their \textit{lack of temporal consistency}: the reward assigned to a given step in the reasoning trajectory is often unrelated to the reward at adjacent steps.
For example, existing works~\cite{lightman2023let, dong2024rlhflow} tend to assign a single scalar value to an entire reasoning trajectory via PRM or ORM, without distinguishing beneficial or suboptimal intermediate steps. 
Meanwhile, models like Generalist Reward Modeling (GRM)~\cite{liu2025dsgrm} often fail to update rewards for current steps by incorporating context from preceding or subsequent steps when generating multi-step reasoning.
This makes it difficult for RMs to distinguish how much each reasoning step contributes to final success, resulting in inconsistent and misleading reward feedback that degrades both \textit{training-time learning signal} (e.g., in RL) and \textit{inference-time search efficiency} (e.g., by encouraging suboptimal trajectories).
These challenges are particularly pronounced in long chain-of-thought (CoT) scenarios (e.g., o1~\cite{qin2024o1}, R1~\cite{ deepseekai2025deepseekr1, yeo2025demystifying}), where models receive no reward until completing a long sequence of reasoning steps.

\begin{figure*}[t!]
    \centering
    \includegraphics[width=1.0\linewidth]{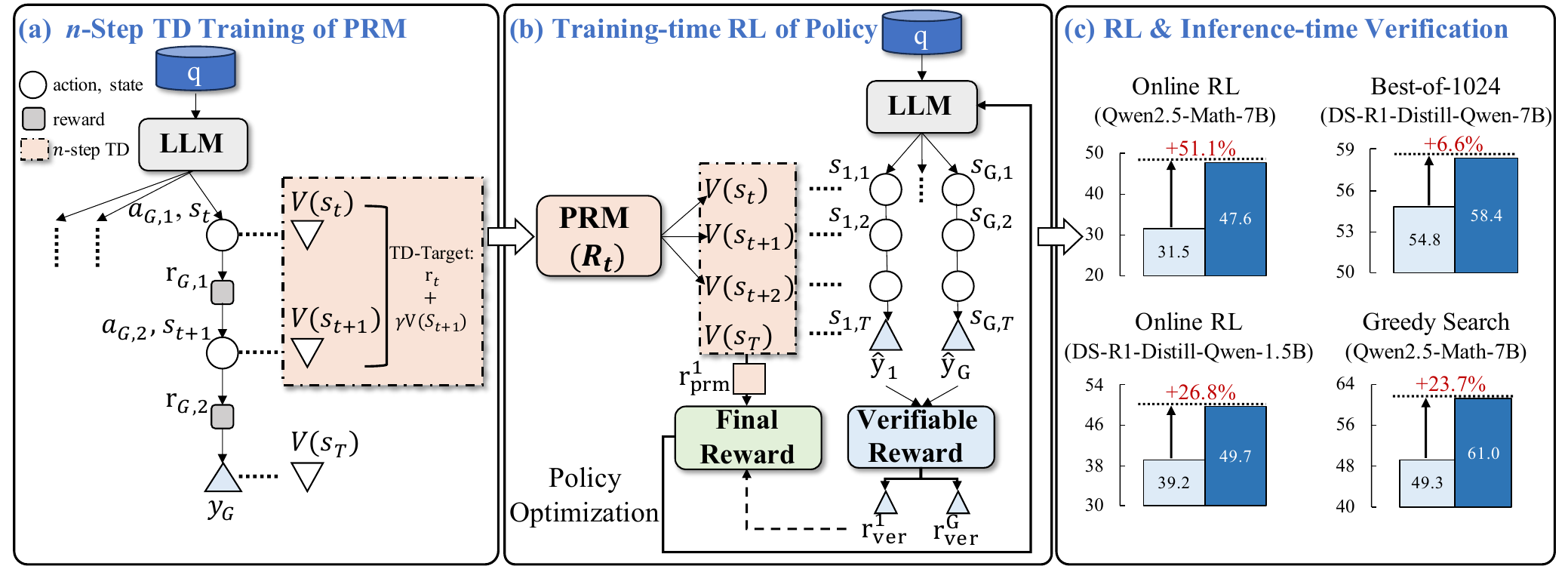}
    \vspace{-0.5cm}
    \caption{
        Overall framework of \model. 
        In panel (a), we employ $n$-step TD learning for training the PRM.
        In panel (b), process reward and verifiable reward are effectively combined for RL training.
        In panel (c), RL training results in Table~\ref{tab: overall_results_rl} compare baselines against \model.
        % using Qwen2.5-Math-7B and DS-R1-Distill-Qwen-1.5B models. 
        Best-of-1024 outcomes in Table~\ref{tab: bon_result_math500} contrast ScalarORM with \model (3-step TD).
        % using a DS-R1-Distill-Qwen-7B backbone. 
        For greedy search evaluations in Figure~\ref{fig:td_search_qwen}, Qwen2.5-Math-7B with a branch factor of 8 is used to compare ScalarORM against \model (TD(2)).
    }
    \vspace{-0.6cm}
    \label{fig: framework}
\end{figure*}

To tackle these challenges, we introduce \model that employs Temporal Difference (TD) learning for reward modeling (Figure~\ref{fig: framework}). 
Unlike prior approaches where TD was used to construct offline datasets, our method leverages TD for online training, dynamically bootstrapping intermediate rewards by integrating future estimates at each step to derive process reward models.
Additionally, we propose a strategy that takes advantage of both rule-based rewards (e.g., from Group Relative Policy Optimization (GRPO)~\cite{shao2024deepseekmath}) and process rewards generated by \model, delivering denser reward signals for online RL training.
We evaluate \model in two scenarios: \textit{(1) inference-time verification} and \textit{(2) training-time reinforcement learning}.
Experimental results show that \model induces \emph{smoother} reward landscapes compared to conventional PRM training --- increasing the low rewards and reducing the high rewards --- thus significantly improving verification accuracy (e.g., Best-of-$N$, tree search) during inference.
\model also demonstrates enhanced RL performance, outperforming multiple LLM baselines in both reward signal density and learning efficiency on mathematical benchmarks.

In summary, our key contributions are listed below:
\begin{itemize}[leftmargin=*,itemsep=0pt,parsep=0.5em,topsep=0.3em,partopsep=0.3em]
    \item We introduce the framework \model, aiming to learn more reliable reward models in RL training. By leveraging temporal difference learning, \model generates smoother reward landscapes in Figure~\ref{fig:reward_distribution}. 
    \item Training-time RL experiments show that incorporating \model into the RL loop yields strong performance gains (up to 51.1\%) and data efficiency (matching 50.1k baseline performance with only 2.5k data) on 8 model variants (5 series) with an effective combination of verifiable rule-based and process rewards in Table~\ref{tab: overall_results_rl}. 
    \item Inference-time verification demonstrates that online TD-trained PRMs significantly enhance performance in both Best-of-$N$ (up to 6.6\%) in Table~\ref{tab: bon_result_math500} and tree-search (up to 23.7\%) in Figure~\ref{fig:td_search_qwen}. 
\end{itemize}

\section{Preliminaries}
\subsection{LLM Reasoning as MDP}
The reasoning process in LLMs can be framed as a Markov Decision Process (MDP)~\cite{tie2025large}.
An MDP typically involves a state space $\mathcal{S}$, containing the full set of possible situations, and an action space $\mathcal{A}$, encompassing the set of allowable decisions. 
It also includes a transition function $f$: $\mathcal{S} \times \mathcal{A}\rightarrow \mathcal{S}$, along with a reward function $R$: $\mathcal{S} \times \mathcal{A} \rightarrow r$, $r \in [0,1]$. 
In our context, the state space corresponds to every possible token sequence generated so far, whereas the action space comprises all possible tokens that can be selected next~\cite{srivastava2025technical}.
The transition function $f$ in our setting is simply the concatenation operation $f(s_t, a_t) = s_t \cdot a_t$, where $\cdot$ denotes concatenation.
Regarding LLM reasoning, the input prompt is given as $(q_0, \cdots, q_L)$, and at step $t-1$, the sequence of generation tokens for a single solution is $(o_0, \cdots, o_{t-1})$. 
Thus, given a prompt, action $a_t$ is a newly generated token and $s_t$ is the token sequence or the context for LLM, i.e., $s_t = (q_0, \cdots, q_L, o_0, \cdots, o_{t-1})$.
In our work, an action is defined as a newly generated sentence.
In standard RL, the reward function $R(s_t, a_t)$ is designed to assign an expected value for the partial generation paths based on each $(s_t, a_t)$ pair. 
In our study, we particularly emphasize establishing a process reward signal at each step $t$ and outcome reward in the terminal step $T$ to guide the judgment (reflecting the correctness of a partial reasoning trace) of generation (guiding the learning direction) of LLM. 
% Reinforcement Learning with Human Feedback (RLHF)~\cite{RLHF} aims to optimize policy $\pi_{\theta}$ via:
% \vspace{-0.2cm}
% \begin{equation}
% \pi^{\star} = \text{arg}\max_{\pi} \mathrm{E}_{{(s_0, \cdots, s_t)}\sim\rho_\pi} \sum_{t=0}^{T} [R(s_t, a_t) - \beta \log \frac{\pi(a_t|s_t)}{\pi_\text{ref}(a_t|s_t)}],
% \end{equation}
% where $\rho_\pi$ is the trajectory distribution, $\pi_\text{ref}$ is the reference model, and $\beta$ controls the distance from the reference model~\cite{lai2024stepdpo, yi2025sppd}.

\subsection{Reward Modeling for LLMs}
\label{ssec: rms}
Recent studies~\cite{lightman2023let, zhang2024rest} model process rewards by training intermediate steps with labeled (Discriminative/Scalar) or generated (Generative) rewards and outcome rewards by comparing the final output with ground truth. 
Specifically, \textit{Process Reward Modeling} estimates the rewards of intermediate steps as hard or soft values using learning a value function or training a value network. 
In contrast, \textit{Rule-based/Outcome Reward Modeling} obtains the outcome reward using a rule-driven function that allocates rewards exclusively according to whether the complete sequence is correct.
In domains such as mathematical reasoning, code generation, and theorem proving, leveraging the final accuracy of verifiable tasks as an outcome reward has proven effective in strengthening reasoning abilities.
Specifically, a correct output will receive a $+1$ reward, while an incorrect output will receive a $0$ reward.
The goal of reward modeling is to help \emph{generalize} to unseen, out-of-distribution (OOD) problems and provide guidance in such OOD scenarios. 

\subsection{Online RL Training}
In this work, we adopt the zero RL training strategy~\cite{zeng2025simplerl} described in DeepSeek-R1~\cite{deepseekai2025deepseekr1}.
This approach utilizes GRPO~\cite{shao2024deepseekmath}, which removes the need for explicit value and advantage functions~\cite{guo2025segment}.
GRPO uses group-normalized rewards to estimate the advantages to further optimize computational efficiency.
For a given query $q$, and the responses $O={o_1, o_2, \dots, o_G}$ are produced by the previous policy model $\pi_\text{old}$. 
The objective of GRPO is to refine the policy model $\pi$ as follows:
\vspace{-0.2cm}
\begin{equation}
\begin{aligned}
\mathcal{J}_\text{GRPO}(\theta)& = \mathbb{E}_{(q,a)\sim \mathcal{D}, \{o_i\}_{i=1}^G\sim \pi_{\theta_\text{old}}(\cdot\mid q)} 
\Bigg[ \frac{1}{G}\sum_{i=1}^{G} \frac{1}{|o_i|}\sum_{j=1}^{|o_i|} \Bigg( 
\min \Big( \frac{\pi_{\theta}(o_{i,j}|q,o_{i,<j})}{\pi_{\theta_\text{old}}(o_{i,j}|q,o_{i,<j})} \hat{A}_{i,j},  
\\& \text{clip} \Big( \frac{\pi_{\theta}(o_{i,j}|q,o_{i,<j})}{\pi_{\theta_\text{old}}(o_{i,j}|q,o_{i,<j})}, 1 - \varepsilon, 1 + \varepsilon \Big) \hat{A}_{i,j} \Big)
- \beta D_{\text{KL}}(\pi_{\theta} || \pi_{\text{ref}}) 
\Bigg) \Bigg],
\label{eq:grpoloss}
\end{aligned}
\end{equation}
where $\pi_\text{ref}$ is the reference model, $o_{i,j}$ represents the token produced at $j$-th generation step in the $i$-th generated response.
To limit deviation from the reference, a KL-divergence regularization term, $D_{\text{KL}}$, is incorporated.
The advantage estimate $\hat{A}_{i,j}$ quantitatively reflects how much each response $o_i$ surpasses the group average. 
This is achieved by normalizing the reward within the group: $\hat{A}_{i,j} = \frac{r_i - \text{mean}(\{r_1, r_2, \dots, r_G\})}{\text{std}(\{r_1, r_2, \dots, r_G\})}$.
The term $r_{i, t}(\theta)$ is defined as the likelihood ratio $\frac{\pi_{\theta}(o_{i,j}|q,o_{i,<j})}{\pi_{\theta_\text{old}}(o_{i,j}|q,o_{i,<j})}$.

\section{The \model Method}
\label{sec: method}
\model employs temporal difference learning to construct reliable reward models for RL training, and can be integrated with verifiable rewards. 
The framework comprises three components (Figure~\ref{fig: framework}):
\begin{itemize}[leftmargin=*,itemsep=0pt,parsep=0.5em,topsep=0.3em,partopsep=0.3em]
    \item PRM Module: A process reward model trained via \textit{n}-step TD learning with reward shaping.
    \item RL Module: Online RL guided by the trained process reward model to optimize policy updates.
    \item \model Integration: An effective linear combination of process reward from PRM and verifiable reward, applied to actor-critic style online RL across different policy model series and sizes.
\end{itemize}

\subsection{Understanding Reward Smoothness}
\label{ssec: reward-smoothness}
\vpara{Background.} Temporal difference (TD) methods enable the iterative refinement of policy value estimates by leveraging the inter-dependencies between states.
In particular, $n$-step TD updates extend this concept by incorporating rewards and value estimates from $n$ subsequent states, providing a more comprehensive and forward-looking perspective compared to traditional $1$-step TD.
This approach discounts future rewards exponentially using a factor (e.g., $\gamma$, usually less than 1) to encourage receiving earlier rewards and balance short-term gains with long-term consequences of actions.

In the context of LLM reasoning, each step corresponds to an individual reasoning operation generated by LLM, and the estimated values serve as process rewards.
We instantiate this framework using the following $n$-step TD algorithm, where $\phi$ represents the parameters of a PRM, to capture the cumulative impact of intermediate reasoning steps and explicitly model the long-term value:
\vspace{-0.2cm}
\begin{equation}
    \phi\leftarrow \phi + 
    \left(\underbrace{\sum_{k=0}^{n-1} \gamma^k r_{t+k} + \gamma^n V(s_{t+n}; \phi)}_\text{TD target} - V(s_t; \phi) \right)
    \cdot \nabla_\phi V(s_t; \phi)
    .
    \label{eq: td_general}
\end{equation}
\vspace{-0.2cm}

\begin{figure}[t!]
    \centering
    \includegraphics[width=1.\linewidth]{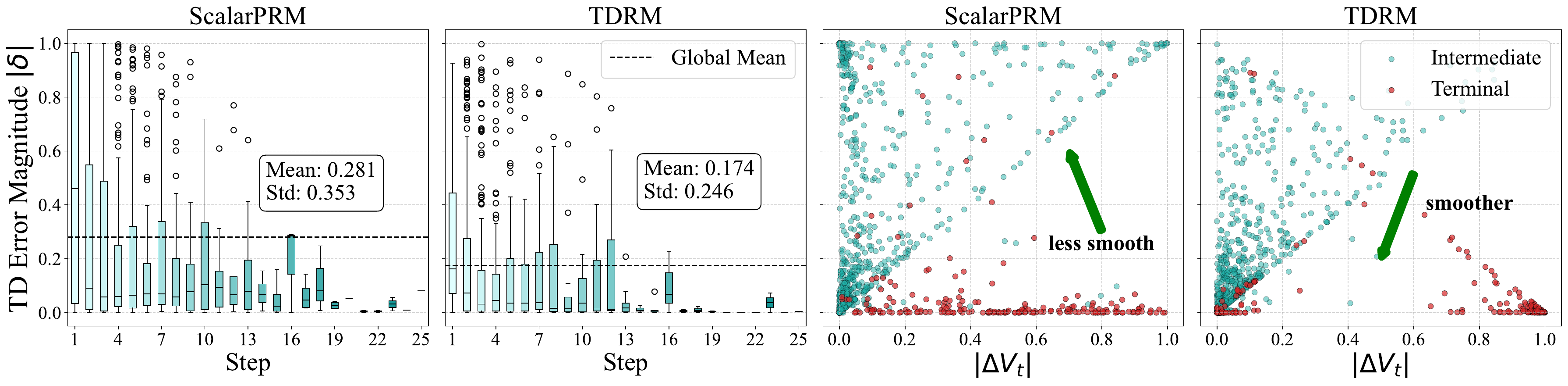}
    \vspace{-0.6cm}
    \caption{Comparison of reward model smoothness. Left: Box plots of TD error magnitude across reasoning steps (steps segmented by double newlines). \model exhibits lower mean and variance of TD errors, indicating smoother and more consistent reward dynamics compared to ScalarPRM. Right: Scatter plots of TD error versus value change magnitude. The tighter distribution in \model shows a more coherent relationship between error and value updates, especially for intermediate steps, while ScalarPRM exhibits noisier and less structured patterns. 
    % \textcolor{red}{(YY: need more details)}
    }
    \label{fig:smoothness_comparison}
\end{figure}

\vpara{Smoothness Analysis.}
Smoothness is a crucial property for effective reward modeling in the reasoning process, as it reflects the consistency and stability of value updates across intermediate steps, ensuring that minor changes in reasoning trajectories do not result in disproportionate deviations in value estimation. 
To measure smoothness, we adopt two complementary approaches to evaluate the behavior of ScalarPRM and our \model. 
\textit{(1) the local Lipschitz constant}, which quantifies the sensitivity of rewards to variations between adjacent states in Table~\ref{tab: Lipschitz_constant} (see details in Appendix~\ref{app: Lipschitz_constant}). 
Our analysis shows that \model yields a smaller Lipschitz constant on average between consecutive steps, indicating smoother reward transitions and better temporal consistency.
\textit{(2) TD error $\delta$} between consecutive reasoning steps and the \textit{value difference $\Delta V_t$} between reasoning steps with Eq.~(\ref{eq: td_error}) and Eq.~(\ref{eq: value_diff}), providing a combined perspective on assessing the continuity and consistency of the estimated value function.
In Figure~\ref{fig:smoothness_comparison}, we compare smoothness by plotting TD error $\delta$ against reasoning steps (steps segmented by double newlines), and TD error $\delta$ vs. value change $\left|\Delta V_t\right|$. 
Here, a step refers to a reasoning segment in the model’s generated trajectory, defined by a double newline delimiter. Examining TD error across steps allows us to assess how consistently the reward model evaluates reasoning as the chain progresses. \model exhibits lower mean and variance of TD errors (0.174 vs. 0.281) than ScalarPRM, indicating smoother and more stable reward dynamics.
In the right panels, each point corresponds to either an intermediate step (\textcolor{cyan}{\textbf{cyan}}, reasoning in progress) or a terminal step (\textcolor{darkred}{\textbf{red}}, final answer). We distinguish between these because terminal steps are evaluated against the final outcome reward, whereas intermediate steps are judged relative to subsequent reasoning states. A smoother relationship between TD error and value changes at intermediate steps indicates that \model provides more coherent trajectory-level reward shaping, while ScalarPRM remains noisier and less structured.
% We observe that \model demonstrates clear temporal consistency: intermediate transitions (\textcolor{cyan}{cyan}) cluster around the origin with error magnitude increasing proportionally with value change, suggesting stable predictions. 
% In contrast, ScalarPRM exhibits a high variance in TD error even when value changes are minimal, indicating less stable value estimation and poor smoothness.
These findings afford insights into a stable and consistent reward model design which motivates our \model.
\begin{table}[h]
    \parbox{0.5\textwidth}{
        \centering
        \caption{Lipschitz constant analysis on average.}
        \vspace{-0.2cm}
        \begin{tabular}{c|c|c}
            \toprule
            Lipz. cont. & ScalarPRM & \model \\
            \midrule
            % Max $\uparrow$ & 1.3673 & 1.6079 \\
            Avg. $\downarrow$ & 0.3331 & 0.2741 \\ 
        \bottomrule
        \end{tabular}
        \label{tab: Lipschitz_constant}
    }
    \parbox{0.48\textwidth}{
        \begin{equation}
            \delta = \left|r + \gamma V(s_{t+1}) - V(s_t)\right| .
            \label{eq: td_error} 
        \end{equation}
        \vspace{-0.2cm}
        \begin{equation}
            \Delta V_t = \left| V(s_{t+1}) - V(s_t) \right|.
            \label{eq: value_diff}
        \end{equation}
        $\delta$ measuring the TD error magnitude and $\Delta V_t$ measuring the value change magnitude.  
    }
\end{table}

\subsection{Reward Modeling}
Motivated by the analyses in Section~\ref{ssec: reward-smoothness} as well as insights from prior research~\cite{yeo2025demystifying} which highlights that the length of CoT does not always increase steadily during LLM reasoning, reward shaping emerges as a crucial mechanism for stabilizing the emergent length scaling behavior.
In the context of our TD-based PRM, reward shaping serves a dual purpose: it refines the TD updates by providing structured feedback and mitigates the volatility of reward signals across different reasoning lengths.

\vpara{Cosine Reward.} 
To stabilize reasoning length, we leverage the cosine-based reward function~\cite{yeo2025demystifying} that adapts to the correctness of reasoning steps and their relative lengths, assigning distinct reward ranges for correct ($Y = 1$) and incorrect ($Y = 0$) steps, formalized as:
\vspace{-0.1cm}
\begin{equation}
r_t =
\begin{cases}
CosRew(L_{gen}, L_{max}, r_0^c, r_L^c), & if \quad Y = 1 \\
CosRew(L_{gen}, L_{max}, r_0^w, r_L^w), & if \quad Y = 0
\end{cases}.
\label{eq: r_t}
\end{equation}
% \vspace{-0.1cm}
Here, $L_{gen}$ represents the current generation length of the reasoning step, while $L_{max}$ denotes the maximum length across all generated steps. 
The parameters $r_0^{c}$ and $r_0^{w}$ specify the initial rewards for correct and incorrect steps when $L_{gen}=0$, set to 1 and 0, respectively. 
Conversely, $r_L^{c}$ and $r_L^{w}$ define the terminal rewards at $L_{gen}=L_{max}$, with values of 2 and -10, respectively.
The binary label $Y$ serves as a correctness indicator for each step.
The cosine reward function itself is defined as:
\vspace{-0.1cm}
\begin{equation}
    CosRew(l, L, r_{min}, r_{max}) = r_{min} + \frac{1}{2}(r_{max} - r_{min})\left(1 + cos\left(\frac{l\pi}{L}\right)\right).
\end{equation}
% \vspace{-0.1cm}
This formulation ensures that the reward begins at its maximum value ($r_{max}$) and gradually decays to the minimum ($r_{min}$) as the reasoning length $l$ approaches the maximum length $L$. 

\vpara{Temporal Difference (TD).} Once the reward function is defined, we can integrate it with the temporal difference framework to update our PRM. 
Leveraging the general TD update formula from Eq.~(\ref{eq: td_general}), we use $r_t$ as the reward in our specific scenario, and set the step size as $1$.
We can then derive our TD target with Eq.~(\ref{eq: td_general}) and Eq.~(\ref{eq: r_t}):
\vspace{-0.1cm}
\begin{equation}
    v_t = \sum_{k=0}^{n-1}\gamma^{k} r_{t+k} + \gamma^{n} V(s_{t+n}).
\end{equation}
% \vspace{-0.1cm}
To align with the desired range of feedback signals, we further process the TD target by clamping it within the interval $[0, 1]$, which yields our final clamped TD target $\tilde{v}_t$:
\vspace{-0.1cm}
\begin{equation}
\tilde{v}_t = 
\left\{
\begin{array}{ll}
    \hfill V_t, & \text{if } \texttt{is\_terminal}(t) \\
    \hfill \min\left(\max\left(\sum_{k=0}^{n-1}\gamma^{k} r_{t+k} + \gamma^{n} V(s_{t+n}), 0\right), 1\right), & \text{otherwise}
\end{array}
\right.
\end{equation}
% \vspace{-0.1cm}
In the terminal states, where no subsequent states exist to contribute to the TD calculation, we directly set the target to $V_t$.
This integration of the custom reward function with TD learning allows our PRM to effectively capture the temporal dynamics of LLM reasoning, providing more informed and stable guidance for policy optimization. 
We present a detailed algorithm in Algorithm \ref{alg:tdn}. 
In Table~\ref{tab:n-step-td}, we set $n$ to each of $\{1,2,3\}$ and explore how different $n$ affects PRM performance.

\vpara{TD-$\lambda$.}
Besides applying $n$-step TD, we also investigate TD-$\lambda$ as an alternative. 
TD-$\lambda$ generalizes $n$-step TD and functions as an online algorithm that offers greater flexibility. 
Due to its online nature, TD-$\lambda$ allows PRM to propagate information to earlier states as soon as it observes a reward. 
For example, in the backward view of TD-$\lambda$, if an intermediate step is incorrect, it can immediately update state values of the preceding states. 
In contrast, in $n$-step TD, the corresponding states would not receive updates until future episodes. 
The pseudo-code and results for PRM training using TD-$\lambda$ are shown in Algorithm~\ref{alg:td_lambda} and Figure~\ref{fig:td-lambda}. 
Notably, in Algorithm~\ref{alg:td_lambda}, we slightly abuse the notation by writing $V(s_t)$ for the value of model logits instead of the sigmoid values.

\vpara{Loss Function.}
In optimizing our PRM, we employ Cross-Entropy Loss that leverages a clamped TD target $\tilde{v}_t$ as a soft label for each reasoning step, enabling the model to learn from the temporal consistency of rewards as:
\vspace{-0.2cm}
\begin{equation}
\mathcal{L}_{\textrm{PRM}} = - \mathbb{E}_{\tau_{\textrm{PRM}} \sim \mathcal{D}_{\textrm{PRM}}} \left[ \frac{1}{|\tau_{\textrm{PRM}}|} \sum_{t=1}^{|\tau_{\textrm{PRM}}|} \tilde{v}_t \log(p_t) + (1 - \tilde{v}_t) \log(1 - p_t) \right],
\end{equation}
where $\tau_{\textrm{PRM}} = \{(s_1, r_1), \dots, (s_T, r_T)\}$ is the trajectory containing each step and the corresponding reward $r_t$, and $p_t$ refers to the model's output probability at step $t$, derived by applying the sigmoid function to the output logits. 
In practice, reasoning steps from diverse trajectories are randomly batched to facilitate minibatch training, ensuring the loss function captures both local step-wise rewards and global trajectory dynamics.

\subsection{Online Reinforcement Learning}
Our algorithm operates online, dynamically calculating TD targets using state values on-the-fly during training. 
Unlike offline algorithms that rely on pre-computed state values, \model adapts to evolving trajectories, leveraging seen trajectories to estimate state values for unseen ones.
This adaptability improves value prediction accuracy and enhances the consistency and robustness of the reward model.

\vpara{Verifiable Reward.}
In our RL training, we follow the verifiable reward $R_{\textrm{verifiable}}$ used in R1~\cite{deepseekai2025deepseekr1}.
$R_{\textrm{verifiable}}$ is defined as a function that checks the format of the predicted answer $\hat{g}$ (\texttt{has\_boxed}) and assesses the equivalence between the prediction $\hat{g}$ and the ground-truth $g$ (\texttt{is\_equivalent}):
\vspace{-0.1cm}
\begin{equation}
    R_{\textrm{verifiable}}(\hat{g}, g) = 
    \begin{cases} 
        1, & \text{if } \texttt{is\_equivalent}(\hat{g}, g) \text{ and } \texttt{has\_boxed}(\hat{g}) \\
        0, & \text{if } \neg\texttt{is\_equivalent}(\hat{g}, g) \text{ and } \texttt{has\_boxed}(\hat{g}) \\
        -1, & \text{otherwise}
    \end{cases}.
    \label{eq: rule}
\end{equation}
% \vspace{-0.1cm}
While $R_{\textrm{verifiable}}$ is straightforward and interpretable, it considers only the end answer and omits assessment of intermediate reasoning steps.
A more detailed explanation of \texttt{is\_equivalent} and \texttt{has\_boxed} can be found in Appendix~\ref{sec: verifiable_reward}.

\vpara{Process-based Reward.} Rule-based verifiable rewards often encounter a critical limitation: they assign identical rewards to trajectories that produce correct answers via incorrect intermediate steps. 
To address this gap and capture the temporal dynamics of reasoning, our PRM plays a pivotal role in online RL. 
By assigning rewards to intermediate states based on their estimated values, PRM provides a more fine-grained feedback signal, effectively mitigating the ``right answer, wrong process'' issue. 
Specifically, the process-based reward at step $t$ is defined as the state value output by the PRM through $R_{\textrm{PRM}}(s_t) := \textrm{PRM}_{\phi}(s_t)$.
% \vspace{-0.1cm}
% \begin{equation}
% R_{\textrm{PRM}}(s_t) := \textrm{PRM}_{\phi}(s_t).
% \label{eq: r_prm}
% \end{equation}
% \vspace{-0.1cm}

\vpara{Effective Combination for RL.}
In \model, we harness the complementary strengths of verifiable and process-based rewards through a linear combination, enabling a more comprehensive and nuanced reward signal for online RL. The final reward function $R_{\textrm{final}}$ is formulated as:
\vspace{-0.1cm}
\begin{equation}
r_{\textrm{final}}=a r_{\textrm{PRM}} + (1-a) r_{\textrm{verifiable}},
\label{eq: r_final}
\end{equation}
% \vspace{-0.1cm}
where the hyper-parameter $a$ balances the influence of process-based feedback against outcome-based verification.
Finally, this combined reward $r_{\textrm{final}}$ as $r_i$ is used in $\hat{A}_{i,j} = \frac{r_i - \text{mean}(\{r_1, r_2, \dots, r_G\})}{\text{std}(\{r_1, r_2, \dots, r_G\})}$ to train the GRPO objective,
% as specified in Eq.~(\ref{eq:grpoloss}), 
enhancing the overall performance and data efficiency of the learning process.

\vpara{Algorithm Implementation.} As presented in Algorithm~\ref{alg:grpo-tdrm}, we outline the overall training process of \model for integrating verifiable and process-based rewards in online RL.  
Additionally, Algorithm~\ref{alg:tdn} provides a step-by-step breakdown of the $n$-step TD method used for training PRM.

% \vspace{-0.1in}
\begin{algorithm}[ht]
\small
\caption{Process of \model}
\label{alg:grpo-tdrm}
\textbf{Notation:} GRPO: group relative policy optimization; PRM$_{\phi}(s_t)$: PRM logits for step $s_t$ \\
\textbf{Input:} Initial policy model $\pi_{\theta}$; process reward model PRM$_\phi$; verifiable reward function $R_{\text{verifiable}}$; task prompts $\mathcal{D}_{\textrm{policy}}$; final reward $R_{\text{final}}$; hyperparameters $\alpha$ \\
\begin{algorithmic}[1]
\STATE Reference model $\pi_{\text{ref}} \leftarrow \pi_{\theta}$
\FOR{Iteration $= 1, \dots, I$}
    \STATE Sample a mini-batch $\mathcal{D}_b$ from $\mathcal{D}_{\textrm{policy}}$
    \STATE Set old policy $\pi_{\text{old}} \leftarrow \pi_{\theta}$
    \STATE Sample $G$ trajectories $\{\tau_i\}_{i=1}^G$ from $\pi_{\text{old}}$ for each question $q \in \mathcal{D}_b$
    \FOR{each trajectory $\tau_i = \{s_1, \dots, s_T\}$}
        \STATE Compute verifiable reward $r_{\text{verifiable}}^{(\tau_i)}$ for $\tau_i$ through Eq.~(\ref{eq: rule}) 
        \STATE Compute process-based reward $r_{\text{PRM}}^{(s_{T-1})} \leftarrow \textrm{PRM}_\phi(s_{T-1})$ for $s_{T-1}$ through $R_{\textrm{PRM}}(s_t) := \textrm{PRM}_{\phi}(s_t)$
        % through Eq.~(\ref{eq: r_prm})
        \STATE Compute final reward $r_{\text{final}}^{(\tau_i)} \leftarrow a \cdot r_{\text{verifiable}}^{(\tau_i)} + (1-a) \cdot r_{\text{PRM}}(s_{T-1})$ for $\tau_i$ through Eq.~(\ref{eq: r_final}) 
    \ENDFOR
    \STATE Compute advantages $\hat{A}_{i,j}$ for the $j$-th token of each $\tau_i$ using group relative advantage estimation
    % \FOR{GRPO iteration $= 1, \dots, \mu$}
    \STATE Update the policy $\pi_\theta$ through maximizing the GRPO objective using $\hat{A}_{i,j}$
    % \ENDFOR
\ENDFOR
\end{algorithmic}
\textbf{Output:} Optimized policy $\pi_\theta$
\end{algorithm}
% \vspace{-0.1in}

\section{Experiments}
\label{sec: exp}
In this section, we benchmark \model in two scenarios, i.e., (1) inference-time verification and (2) training-time online reinforcement learning.

\subsection{Experimental settings}
\label{ssec: exp_setting}
\vpara{Evaluation Metrics and Benchmarks.} 
\textit{(1) For inference-time verification}, we compare different reward models under two key settings.
\textit{Best-of-$N$ Sampling} works by first generating a pool of $N$ potential outputs and selecting the best candidate using the RM.
We test with $N \in \{128, 1024\}$ and evaluate on GSM8K~\cite{cobbe2021training} and MATH-500~\cite{hendrycks2021measuring}.
\textit{Greedy Search}~\cite{light2025disc} generates outputs by iteratively selecting the highest-scoring sequences. 
To improve exploration, the branching factor is set to $m \in \{2, 4, 8, 16\}$, and experiments are performed on MATH-500.
For a fair comparison, pre-generate reasoning trajectories are utilized during inference.
Accuracy is used as the evaluation metric for both strategies (see more details in Appendix~\ref{appendix: experiment}).
\textit{(2) For training-time online RL}, we benchmark \model against leading methods on five difficult datasets: MATH-500, Minerva Math~\cite{lewkowycz2022solvingMinervaMath}, Olympiad Bench~\cite{he2024olympiadbench}, AIME24, and AMC23. 
Following SimpleRL~\cite{zeng2025simplerl}, we evaluate performance using the Pass@1 metric with greedy decoding.
\begin{wrapfigure}{r}{0.45\textwidth}
    \centering
    \includegraphics[width=\linewidth]{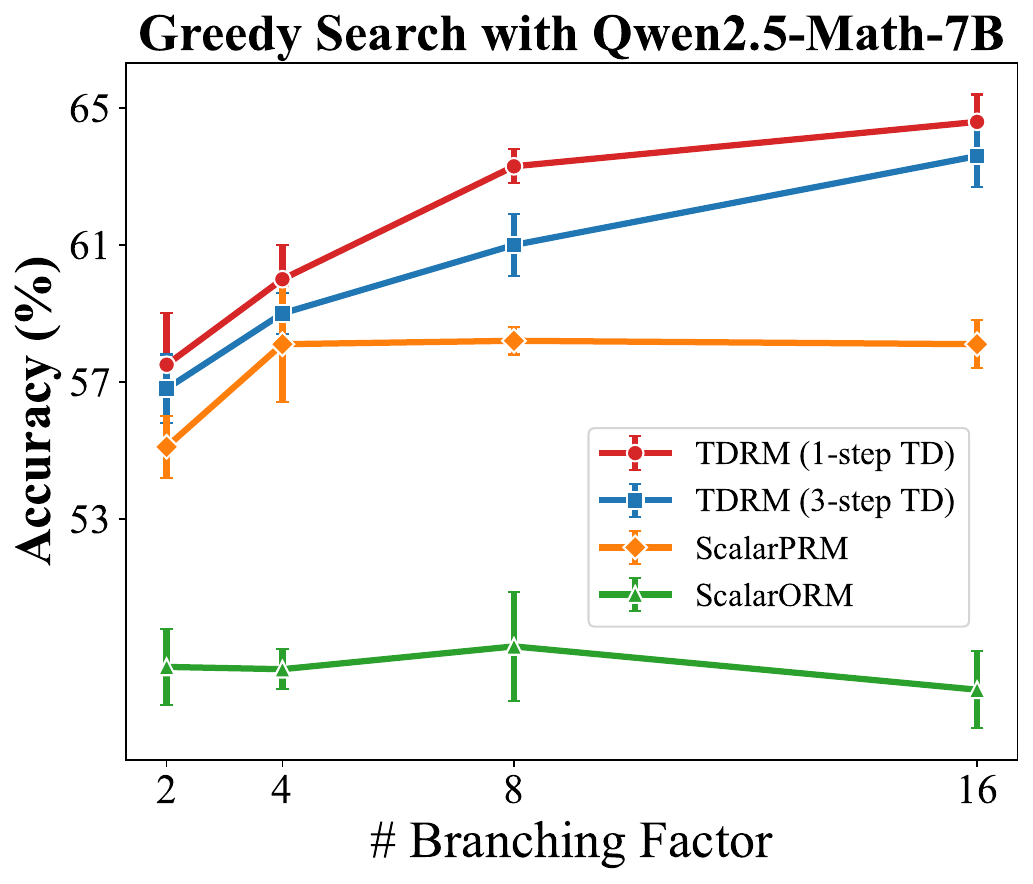}
    \caption{Comparison of \model versus baselines on Greedy Search, using Qwen2.5-Math-7B as the backbone.}
    \label{fig:td_search_qwen}
    \vspace{-3.5em}
\end{wrapfigure}

\subsection{Main Experimental Results}
\begin{table}[t!]
    \centering
    \parbox{0.71\textwidth}{
        \centering
        \caption{Results on MATH-500 using the RM for selection in Best-of-$N$ sampling. The PRM backbone is DeepSeek-R1-Distill-Qwen-7B.}
        \vspace{-0.2cm}
        \resizebox{!}{12mm}{
        \begin{tabular}{c|c|c|c|c}   
        \specialrule{.16em}{0pt}{.65ex}   
        \multirow{2}{*}{Method} & \multicolumn{2}{c|}{DS-R1-Distill-Qwen-7B} & \multicolumn{2}{c}{Llama3.1-8B-Instruct} \\   \cmidrule(lr){2-5}   & Best-of-128 & Best-of-1024 & Best-of-128 & Best-of-1024 \\   
        \specialrule{.10em}{.4ex}{.65ex} 
        ScalarORM & 52.0 & 54.8 & 42.2 & 42.8 \\   
        ScalarPRM & 53.4 & 56.2 & \textbf{44.4} & 44.8 \\   
        \model & \textbf{54.2} & \textbf{58.4} & 43.2 & \textbf{45.6} \\   
        \specialrule{.16em}{.4ex}{0pt}
        \end{tabular}
        }
    \label{tab: bon_result_math500}
    }
    ~
    \parbox{0.23\textwidth}{
        \centering
        \caption{Results on GSM8K in Best-of-128 sampling.}
        \vspace{-0.2cm}
        \resizebox{!}{10mm}{
        \begin{tabular}{c|c}   
            \specialrule{.16em}{0pt}{.65ex}   
            Method & Result \\   
            \specialrule{.10em}{.4ex}{.65ex}   
            ScalarORM & 69.29\\   
            ScalarPRM & 71.34\\   
            \model & \textbf{73.24} \\   
            \specialrule{.16em}{.4ex}{0pt}
        \end{tabular}
        }
        \label{tab: bon_result_gsm8k}
    }
\end{table}

% \begin{figure*}[t!]
%     \centering
%     \begin{subfigure}[b]{0.46\textwidth}
%         \centering
%         \label{fig: td-search-qwen}
%         \includegraphics[height=2.2in]{figs/tree_search_qwen.pdf}
%         % \caption{}
%     \end{subfigure}
%     ~
%     \begin{subfigure}[b]{0.46\textwidth}
%         \centering
%         \label{fig: td-search-mistral}
%         \includegraphics[height=2.2in]{figs/tree_search_mistral.pdf}
%         % \caption{}
%     \end{subfigure}
%     \vspace{-0.4cm}
%     \caption{Results of greedy search on our PRM with TD.}
%     \label{fig:td_search}
% \end{figure*}

\begin{table}[t!]
    \centering
    \caption{Evaluation results on standard mathematical benchmarks under a constrained data system of 2.5k samples. We highlight the top score in \textbf{bold} and the second-best by \underline{underlining} it. The relative improvement (\textcolor{darkred}{\textbf{\%Improv.}}) for each method is computed based on the performance in this setup.}
    \vspace{-0.2cm}
    \label{tab: overall_results_rl}
    \resizebox{0.9\textwidth}{!}{  
        \begin{tabular}{lllllllc}
            \specialrule{.16em}{0pt}{.65ex}  
            Model & \begin{tabular}[c]{@{}l@{}}Data \\ Size
            \end{tabular}                           & \begin{tabular}[c]{@{}l@{}}MATH \\ 500\end{tabular}                     & \begin{tabular}[c]{@{}l@{}}Minerva \\
            Math\end{tabular} & \begin{tabular}[c]{@{}l@{}}Olympiad \\ Bench\end{tabular} & \begin{tabular}[c]{@{}l@{}}AIME24 \\ (Pass@1)\end{tabular} & AMC23                        & Avg.       \\ 
            \specialrule{.10em}{.4ex}{.65ex}  
            \multicolumn{8}{c}{\textbf{Backbone is Base Model, Qwen Series}} \\
            \specialrule{.10em}{.4ex}{.65ex}  
            \multicolumn{1}{l}{\textbf{Qwen2.5-0.5B}}        & \multicolumn{1}{l}{-}         &15.8	&4.8	&2.8	&0.0	&12.5	&7.2         \\ 
            + SimpleRL (Greedy)             & 50.1k                     	&32.6	&8.1	&9.0	&0.0	&15.0	&12.9 \\
            \specialrule{.10em}{.4ex}{.65ex}  
            + ScalarPRM    & \multirow{4}{*}{2.5k} &3.4	&2.2	&1.9	&0.0	&5.0	&2.5 \\
            + ScalarORM    &  &6.2	&2.2	&2.8	&0.0	&5.0	&3.2  \\
            + Rule-based    &  	&29.8	&4.0	&7.0	&0.0	&12.5	&\underline{10.7} \\ 
            \rowcolor{table-blue!66}  + Ours             &  	&26.2	&4.8	&7.1	&0.0	&15.0	&\textbf{10.8} (\textbf{\textcolor{darkred}{+0.9\%}}) \\
            \specialrule{.10em}{.4ex}{.65ex}
            \multicolumn{1}{l}{\textbf{Qwen2.5-1.5B}}        & \multicolumn{1}{l}{-}         &29.6	&6.6	&6.5	&0.0	&12.5	&11.0  \\ 
            + SimpleRL (Greedy) & 50.1k   	&22.6	&2.6	&8.4	&0.0	&5.0	&7.7  \\    
            \specialrule{.10em}{.4ex}{.65ex} 
            + ScalarPRM    & \multirow{4}{*}{2.5k}	&0.2	&0.0	&0.0	&0.0	&0.0	&0.0 \\ 
            + ScalarORM    & 	&1.8	&0.0	&0.6	&0.0	&5.0	&1.5 \\ 
            + Rule-based    &  &58.0	&12.1	&18.7	&0.0	&27.5 &\underline{23.3} \\ 
            \rowcolor{table-blue!66}  + Ours             &  	&52.8	&9.9	&17.8	&3.3	&35.0 &\textbf{23.8} (\textbf{\textcolor{darkred}{+2.1\%}}) \\
            \specialrule{.10em}{.4ex}{.65ex}  
            \multicolumn{8}{c}{\textbf{Backbone is Chat Model, GLM Series}}\\
            \specialrule{.10em}{.4ex}{.65ex} 
            \textbf{GLM4-9B-0414} &- &65.8	&36.8	&28.7	&10.0	&42.5  &36.8 \\
            \specialrule{.10em}{.4ex}{.65ex} 
            + ScalarPRM    &\multirow{4}{*}{2.5k} &67.0	&38.6	&31.9	&6.7	&45.0 & 37.8\\
            + ScalarORM  & &68.2	&39.3	&30.2	&10.0	&42.5 &38.0 \\    
            + Rule-based & &72.8	&37.5	&37.0	&16.7	&40.0  &\underline{40.8} \\ 
            \rowcolor{table-blue!66} + Ours    & &72.2	&37.1	&32.0	&20.0	&47.5	&\textbf{41.8} (\textbf{\textcolor{darkred}{+2.5\%}}) \\
            \specialrule{.10em}{.4ex}{.65ex}  
            \multicolumn{8}{c}{\textbf{Backbone is Reasoning Model, GLM Series}}\\
            \specialrule{.10em}{.4ex}{.65ex} 
            \textbf{GLM-Z1-9B-0414} &- &93.6	&43.8	&65.5	&73.3	&92.5 &73.7 \\
            \specialrule{.10em}{.4ex}{.65ex} 
            + ScalarPRM    &\multirow{4}{*}{2.5k} &94.0	&47.8	&66.4	&76.7	&92.5 &75.5\\
            + ScalarORM  & &95.0	&46.7	&65.9	&76.7	&97.5 &\underline{76.4} \\    
            + Rule-based & &95.6	&43.4	&65.2	&73.3	&97.5 &75.0 \\ 
            \rowcolor{table-blue!66} + Ours    & &94.6	&44.9	&66.5	&80.0	&97.5  &\textbf{76.7} (\textbf{\textcolor{darkred}{+0.4\%}}) \\
            \specialrule{.10em}{.4ex}{.65ex}  
            \multicolumn{8}{c}{\textbf{Backbone is Base Model, Qwen-Math Series}} \\
            \specialrule{.10em}{.4ex}{.65ex}  
            \multicolumn{1}{l}{\textbf{Qwen2.5-Math-1.5B}}        & \multicolumn{1}{l}{-}         &42.2	&8.8	&27.0	&10.0	&37.5 &25.1  \\ 
            + SimpleRL (Greedy) & 50.1k   	&59.8 & 13.6 &29.9 &10.0 &37.5 &30.2  \\            
            \specialrule{.10em}{.4ex}{.65ex} 
            + ScalarPRM    & \multirow{4}{*}{2.5k}	&66.2	&17.3	&28.7	&13.3	&50.0 &35.1 \\ 
            + ScalarORM    & 	&41.6	&8.5	&27.0	&
            10.0	&40.0 &25.4 \\ 
            + Rule-based    &  &67.6	&21.3	&31.0	&6.7	&52.5 &\underline{35.8} \\
            \rowcolor{table-blue!66}  + Ours             &  	&66.2	&18.4	&30.1	&13.3	&55.0 &\textbf{36.6} (\textbf{\textcolor{darkred}{+2.2\%}}) \\
            \specialrule{.10em}{.4ex}{.65ex}
            \multicolumn{1}{l}{\textbf{Qwen2.5-Math-7B}}        & \multicolumn{1}{l}{-}           & {63.6}  & {12.5}    & {25.8}      & {13.3}       & {42.5}  & 31.5         \\ 
            + Our Template & \multicolumn{1}{l}{-}         & {68.8}  & {16.2}    & {31.1}      & {13.3}       & {62.5}  & 38.4         \\ 
            + SimpleRL-Zero & 8.5k                       & 77.8  & 31.2    & 37.5        & 23.3         & 62.5    & 46.5 \\
            + SimpleRL (Greedy)             & 50.1k                        & {78.2}  & {27.6}    & {40.3}      & {26.7}       & {60.2}  & {46.6}\\
            GRPO & - & 77.8	&39.7	&39.1	&20.0	&57.5		&46.8 \\ 
            Dr. GRPO† & - & 74.6	&30.1	&37.3	&26.7	&50.0		&43.7 \\
            OpenReasoner-Zero &- & 82.4	&31.6	&47.9	&13.3	&54.2		&45.9 \\
            \specialrule{.10em}{.4ex}{.65ex}  
            + ScalarPRM    & \multirow{4}{*}{2.5k} & 75.8& 29.0               & 36.4               & 26.7                & 60.0  & \underline{45.6}\\
            + ScalarORM    & & 71.2& 22.1             & 37.5               & 20.0                  & 50.0  & 40.2         \\
            + Rule-based            &  & 73.2& 25.0               & 37.8               & 23.3                & 65.0  & 44.9 \\
            \rowcolor{table-blue!66}  + Ours             &  & 74.6 & 26.8   & 37.3     & 36.7      & 62.5 & \textbf{47.6} (\textbf{\textcolor{darkred}{+4.4\%}}) \\
            \specialrule{.10em}{.4ex}{.65ex} 
            \multicolumn{8}{c}{\textbf{Backbone is Reasoning Model, DeepSeek Series}}\\
            \specialrule{.10em}{.4ex}{.65ex}  
            \textbf{DS-R1-Distill-Qwen-1.5B} &- &70.6	&26.5	&32.1	&16.7	&50.0  &39.2 \\
            \specialrule{.10em}{.4ex}{.65ex} 
            + ScalarPRM &\multirow{4}{*}{2.5k} &74.2	&29.0	&35.7	&33.3	&60.0 &46.4 \\
            + ScalarORM & &77.4	&30.5	&38.5	&33.3	&60.0		&\underline{47.9} \\
            + Rule-based & &75.4	&26.8	&36.1	&20.0	&57.5 &43.2 \\
            \rowcolor{table-blue!66} + Ours & &79.8	&30.5	&38.2	&30.0	&70.0  &\textbf{49.7} (\textbf{\textcolor{darkred}{+3.8\%}}) \\
            \specialrule{.10em}{.4ex}{.65ex}  
            \textbf{DS-R1-Distill-Qwen-7B} &- &88.0	&43.0	&49.9	&63.3	&82.5  &65.3 \\
            SEED-GRPO &8.5k &91.6	&38.6	&61.5	&50.0	&78.3  &64.0 \\
            \specialrule{.10em}{.4ex}{.65ex} 
            + ScalarPRM    &\multirow{4}{*}{2.5k} &87.6	&50.7	&49.8	&53.3	&85.0  &65.3 \\ 
            + ScalarORM  & &90.4	&50.7	&52.7	&43.3	&90.0  &\underline{65.4} \\   
            + Rule-based & &89.6	&46.0	&52.4	&50.0	&82.5  &64.1 \\
            \rowcolor{table-blue!66} + Ours    & &91.8	&50.4	&54.1	&53.3	&87.5  &\textbf{67.4} (\textbf{\textcolor{darkred}{+3.0\%}}) \\
             \specialrule{.16em}{.4ex}{0pt}
        \end{tabular}
    }
\end{table}

\vpara{Reward Modeling and Inference Scaling Results.}
Corresponding Section~\ref{ssec: rms}, Figure~\ref{fig: RMs} provides the definition comparison and MATH-500 results of recent reward models.
Table~\ref{tab: bon_result_math500} and \ref{tab: bon_result_gsm8k} present the results of Best-of-$N$ sampling across different models and datasets, providing empirical evidence of \model's superiority. 
Firstly, \model outperforms ScalarPRM and ScalarORM on the MATH-500 dataset as the sampling budget increases from Best-of-128 to Best-of-1024.  
Specifically, with DS-R1-Distill-Qwen-7B, \model achieves relative improvements of 6.7\% over ScalarORM and 3.9\% over ScalarPRM; with Llama3.1-8B-Instruct, the respective relative gains are 6.5\% compared to ScalarORM and 1.8\% compared to ScalarPRM.
This strongly indicates that \model is more reliable and can consistently identify the best responses with larger sampling budgets. 
Notably, the GSM8K results in Table~\ref{tab: bon_result_gsm8k} utilize samples generated by \texttt{Mistral-7B-Instruct-v0.2}, which is different from the reward models' training data, demonstrating \model's superior ability to generalize to new data distributions.  
%As an OOD experiment, it further validates the \model's ability to generalize, as it maintains higher accuracy compared to baseline methods when assessing unseen data distributions.

In tree search evaluations, as shown in Figure~\ref{fig:td_search_qwen}, \model again demonstrates superior performance with Qwen2.5-Math-7B and provides a more accurate verification of reasoning trajectories.
Moreover, \model exhibits enhanced reliability, with its accuracy improving as the number of search branching factors increases from 2 to 16, indicating its effectiveness in navigating complex decision spaces. 
In addition, as shown in Figure~\ref{fig:td_search_mistral}, \model further validates its ability on unseen data distributions (i.e., Mistral data) compared to baseline methods.

\vpara{Online Reinforcement Learning Results.}
Table~\ref{tab: overall_results_rl} compares the RL training outcomes of \model against the state-of-the-art methods, demonstrating its superiority across 8 model variants (5 series) using only 2.5k MATH Level-3 prompts.
Spanning diverse model sizes and pre-training paradigms, \model consistently achieves the highest average accuracy, underscoring its reliability in RL training.
For example, \model beats all the other methods --- whether using verifiable rewards or reward models --- by an average of $0.9\%$ to $4.4\%$ over the second-best model, with a notable $36.7\%$ Pass@1 on AIME24 on Qwen2.5-Math-7B, highlighting its significant advancement in mathematical reasoning. 
Notably, on smaller Qwen2.5-(0.5B, 1.5B), which exhibit weaker inherent math capabilities, training with ScalarPRM or ScalarORM alone leads to model collapse.
In contrast, \model’s linear combination of verifiable and process-based rewards ensures stable performance and superior data efficiency, enabling consistent learning even with limited training samples.

\noindent
\begin{minipage}[t!]{0.32\textwidth}
    \centering
    % \vspace{-2.7cm} % Manual alignment fudge factor; adjust if needed
    \captionsetup{type=table}
    \captionof{table}{Results of $n$-step TD on MATH-500 when backbone is DS-R1-Distill-Qwen-7B.}
    \vspace{-0.2cm}
    \label{tab:n-step-td}
    \resizebox{\textwidth}{!}{
    \begin{tabular}{c|c|c}
        \specialrule{.16em}{0pt}{.65ex}  
        $n$ & Best-of-128 & Best-of-1024 \\
        \specialrule{.10em}{.4ex}{.65ex}   
        1 & 54.2 & \textbf{58.4} \\
        2 & \textbf{55.4} & 56.2 \\
        3 & 54.2 & 56.8 \\
        \specialrule{.16em}{.4ex}{0pt}
    \end{tabular}
    }
\end{minipage}
\hfill
\begin{minipage}[t!]{0.3\textwidth}
    \centering
    \includegraphics[width=0.95\linewidth]{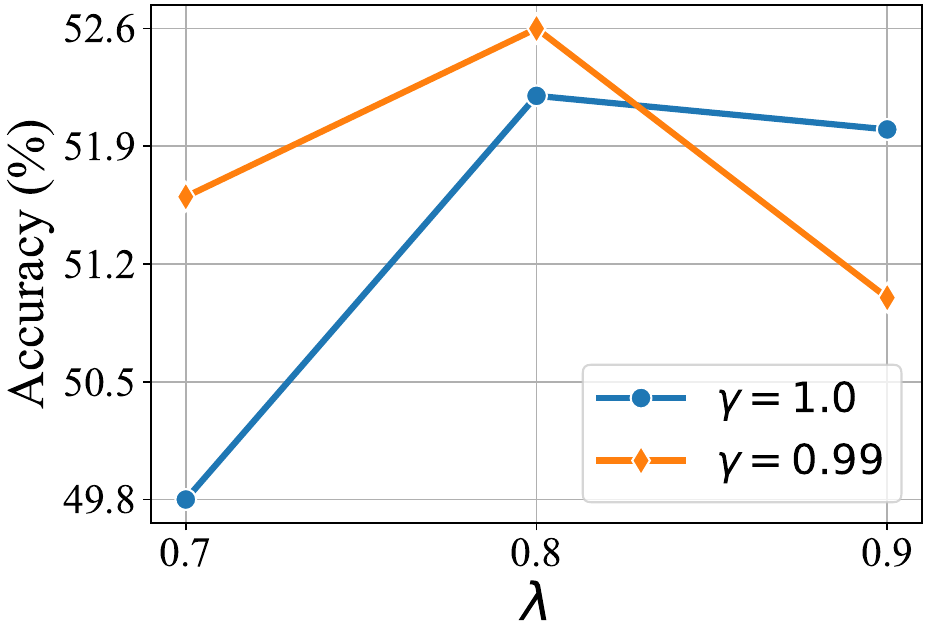}
    \vspace{-0.2cm}
    % \captionof{figure}{Results of Best-of-128 on MATH-500 vs TD-$\lambda$.}
    \captionof{figure}{TD-$\lambda$ results on MATH-500 in Best-of-128 sampling.}
    \label{fig:td-lambda}
\end{minipage}
\hfill
\begin{minipage}[t!]{0.3\textwidth}
    \centering
    \includegraphics[width=0.95\linewidth]{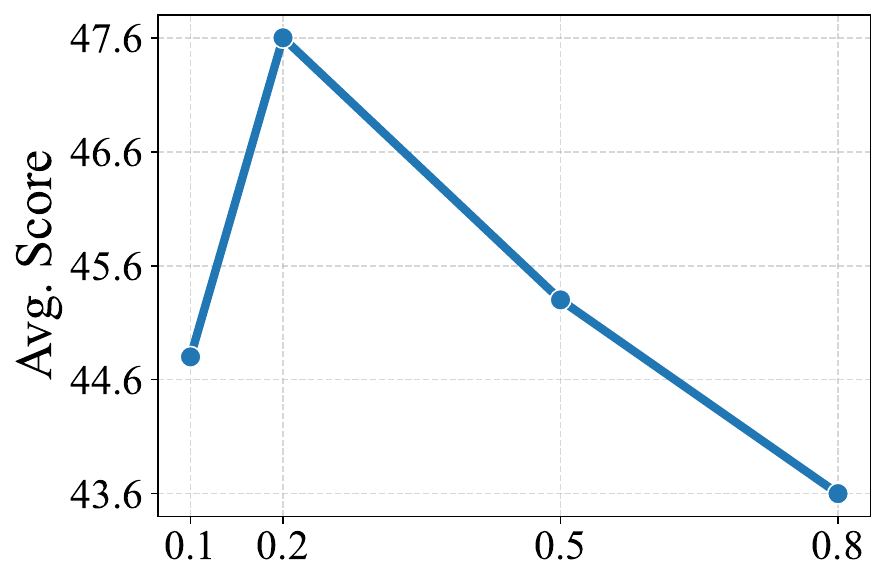}
    \vspace{-0.2cm}
    \captionof{figure}{Avg. performance of online RL training vs $a$ on Qwen2.5-Math-7B.}
    \label{fig:alpha-sweep}
\end{minipage}

\subsection{Analysis and ablation studies}
We study the features of \model through comprehensive analyses: reward distribution comparison, varying lookahead steps in \model, TD-$\lambda$, and the tradeoff between verifiable and process rewards.
Given that \model (3-step TD) is the primary configuration for RL training, the following studies focus on this setup, with ScalarPRM serving as the default comparator unless specified otherwise.

\vpara{\textit{n}-step TD.}
To study the effect of look-ahead steps $n$, we present the Best-of-$N$ results for \model trained with 1, 2, and 3-step TD in Table~\ref{tab:n-step-td}.
While 1-step TD performs the best under a larger sampling budget (Best-of-1024), 2-step TD achieves the best under a smaller number of candidates (Best-of-128). 
This suggests that a moderate lookahead step may help improve sample efficiency, while shorter horizons exhibit greater robustness and generalize better with a larger budget.

\vpara{TD-$\lambda$.} Building on the $n$-step TD framework, where TD-$\lambda$ provides a mechanism to balance between TD(0) and TD(1), we evaluate TD-$\lambda$ under varying values of $\lambda$ and different  discount factor $\gamma$.% in Best-of-128 sampling.
As shown in Figure~\ref{fig:td-lambda}, the interaction between $\lambda$ and $\gamma$ has a significant non-linear impact on model accuracy.
Specifically, $\lambda = 0.8$ consistently achieves the highest accuracy for both discount factors. 
However, as $\lambda$ increases to $0.9$, the accuracy declines.
These results highlight that tuning of $\lambda$ around $0.8$ is critical for balancing temporal consistency and achieving optimal performance.

\vpara{Reward Combination Tradeoff.}
In the RL training of \model, we analyze the linear combination of the process reward and the verifiable reward via the coefficient $a$ (verifiable reward coefficient: $1-a$).
As shown in Figure~\ref{fig:alpha-sweep}, performance peaks at $a=0.2$, with significant degradation for both higher and lower values.
This indicates process rewards serve best as a complementary signal---too low a weight introduces insufficient guidance, while excessive weight amplifies noise. 

\section{Conclusion}
\model tackles the challenge of temporal inconsistency in reward models by introducing TD regularization, which enhances reward density and stability. Across Best-of-$N$ and tree-search scenarios, \model-trained PRMs consistently improve performance and complement verifiable reward methods, enabling more data-efficient RL training and stronger LLM policies on 8 models.  

These results show that incorporating temporal consistency into reward models not only stabilizes RL training but also opens the door to more scalable RLHF pipelines, higher-quality inference-time search, and broader applications in aligning LLMs with complex objectives.

% Our approach paves the way for more reliable reward modeling with broader implications to bolster the stability and efficacy of LLM-based RL frameworks.

% \clearpage

\subsubsection*{Reproducibility Statement}
\label{sec: reproducibility}
We provide pseudo-code in the Algorithm~\ref{alg:grpo-tdrm}, Algorithm~\ref{alg:td_lambda}, and Algorithm~\ref{alg:tdn} for \model training process, TD-$\lambda$, and $n$-step TD for PRM training.
We provide experimental settings in Section~\ref{ssec: exp_setting} and \ref{appendix: experiment}.
We also release all code to promote reproducibility.
% We also release all code to promote reproducibility at \url{https://anonymous.4open.science/r/TDRM-CDD6}.

\bibliography{ref}

\begin{thebibliography}{48}
\providecommand{\natexlab}[1]{#1}
\providecommand{\url}[1]{\texttt{#1}}
\expandafter\ifx\csname urlstyle\endcsname\relax
  \providecommand{\doi}[1]{doi: #1}\else
  \providecommand{\doi}{doi: \begingroup \urlstyle{rm}\Url}\fi

\bibitem[Achiam et~al.(2023)Achiam, Adler, Agarwal, Ahmad, Akkaya, Aleman, Almeida, Altenschmidt, Altman, Anadkat, et~al.]{achiam2023gpt}
Josh Achiam, Steven Adler, Sandhini Agarwal, Lama Ahmad, Ilge Akkaya, Florencia~Leoni Aleman, Diogo Almeida, Janko Altenschmidt, Sam Altman, Shyamal Anadkat, et~al.
\newblock Gpt-4 technical report.
\newblock \emph{arXiv preprint arXiv:2303.08774}, 2023.

\bibitem[Ahmadian et~al.(2024)Ahmadian, Cremer, Gall{\'e}, Fadaee, Kreutzer, Pietquin, {\"U}st{\"u}n, and Hooker]{ahmadian2024rloo}
Arash Ahmadian, Chris Cremer, Matthias Gall{\'e}, Marzieh Fadaee, Julia Kreutzer, Olivier Pietquin, Ahmet {\"U}st{\"u}n, and Sara Hooker.
\newblock Back to basics: Revisiting reinforce style optimization for learning from human feedback in llms.
\newblock \emph{arXiv preprint arXiv:2402.14740}, 2024.

\bibitem[Cheng et~al.(2024)Cheng, Liu, Wang, Gu, Lu, Zhang, Dong, Tang, Wang, and Huang]{cheng2024spar}
Jiale Cheng, Xiao Liu, Cunxiang Wang, Xiaotao Gu, Yida Lu, Dan Zhang, Yuxiao Dong, Jie Tang, Hongning Wang, and Minlie Huang.
\newblock Spar: Self-play with tree-search refinement to improve instruction-following in large language models.
\newblock \emph{arXiv preprint arXiv:2412.11605}, 2024.

\bibitem[Cobbe et~al.(2021)Cobbe, Kosaraju, Bavarian, Chen, Jun, Kaiser, Plappert, Tworek, Hilton, Nakano, et~al.]{cobbe2021training}
Karl Cobbe, Vineet Kosaraju, Mohammad Bavarian, Mark Chen, Heewoo Jun, Lukasz Kaiser, Matthias Plappert, Jerry Tworek, Jacob Hilton, Reiichiro Nakano, et~al.
\newblock Training verifiers to solve math word problems.
\newblock \emph{arXiv preprint arXiv:2110.14168}, 2021.

\bibitem[DeepSeek(2025)]{deepseekai2025deepseekr1}
DeepSeek.
\newblock Deepseek-r1: Incentivizing reasoning capability in llms via reinforcement learning.
\newblock \emph{arXiv preprint arXiv:2501.12948}, 2025.

\bibitem[Dong et~al.(2024)Dong, Xiong, Pang, Wang, Zhao, Zhou, Jiang, Sahoo, Xiong, and Zhang]{dong2024rlhflow}
Hanze Dong, Wei Xiong, Bo~Pang, Haoxiang Wang, Han Zhao, Yingbo Zhou, Nan Jiang, Doyen Sahoo, Caiming Xiong, and Tong Zhang.
\newblock Rlhf workflow: From reward modeling to online rlhf.
\newblock \emph{arXiv preprint arXiv:2405.07863}, 2024.

\bibitem[GLM et~al.(2024)GLM, Zeng, Xu, Wang, Zhang, Yin, Rojas, Feng, Zhao, Lai, et~al.]{glm2024chatglm}
Team GLM, Aohan Zeng, Bin Xu, Bowen Wang, Chenhui Zhang, Da~Yin, Diego Rojas, Guanyu Feng, Hanlin Zhao, Hanyu Lai, et~al.
\newblock Chatglm: A family of large language models from glm-130b to glm-4 all tools.
\newblock \emph{arXiv preprint arXiv:2406.12793}, 2024.

\bibitem[Guo et~al.(2025)Guo, Xu, Liu, Ye, and Qiu]{guo2025segment}
Yiran Guo, Lijie Xu, Jie Liu, Dan Ye, and Shuang Qiu.
\newblock Segment policy optimization: Effective segment-level credit assignment in rl for large language models.
\newblock \emph{arXiv preprint arXiv:2505.23564}, 2025.

\bibitem[He et~al.(2024)He, Luo, Bai, Hu, Thai, Shen, Hu, Han, Huang, Zhang, et~al.]{he2024olympiadbench}
Chaoqun He, Renjie Luo, Yuzhuo Bai, Shengding Hu, Zhen~Leng Thai, Junhao Shen, Jinyi Hu, Xu~Han, Yujie Huang, Yuxiang Zhang, et~al.
\newblock Olympiadbench: A challenging benchmark for promoting agi with olympiad-level bilingual multimodal scientific problems.
\newblock In \emph{ACL}, pp.\  3828--3850, 2024.

\bibitem[Hendrycks et~al.(2021)Hendrycks, Burns, Kadavath, Arora, Basart, Tang, Song, and Steinhardt]{hendrycks2021measuring}
Dan Hendrycks, Collin Burns, Saurav Kadavath, Akul Arora, Steven Basart, Eric Tang, Dawn Song, and Jacob Steinhardt.
\newblock Measuring mathematical problem solving with the math dataset.
\newblock \emph{arXiv preprint arXiv:2103.03874}, 2021.

\bibitem[Hu et~al.(2025)Hu, Zhang, Han, Jiang, Zhang, and Shum]{hu2025open}
Jingcheng Hu, Yinmin Zhang, Qi~Han, Daxin Jiang, Xiangyu Zhang, and Heung-Yeung Shum.
\newblock Open-reasoner-zero: An open source approach to scaling up reinforcement learning on the base model.
\newblock \emph{arXiv preprint arXiv:2503.24290}, 2025.

\bibitem[Jaech et~al.(2024)Jaech, Kalai, Lerer, Richardson, El-Kishky, Low, Helyar, Madry, Beutel, Carney, et~al.]{jaech2024openai}
Aaron Jaech, Adam Kalai, Adam Lerer, Adam Richardson, Ahmed El-Kishky, Aiden Low, Alec Helyar, Aleksander Madry, Alex Beutel, Alex Carney, et~al.
\newblock Openai o1 system card.
\newblock \emph{arXiv preprint arXiv:2412.16720}, 2024.

\bibitem[Jiang et~al.(2023)Jiang, Sablayrolles, Mensch, Bamford, Chaplot, de~las Casas, Bressand, Lengyel, Lample, Saulnier, Lavaud, Lachaux, Stock, Scao, Lavril, Wang, Lacroix, and Sayed]{jiang2023mistral7b}
Albert~Q. Jiang, Alexandre Sablayrolles, Arthur Mensch, Chris Bamford, Devendra~Singh Chaplot, Diego de~las Casas, Florian Bressand, Gianna Lengyel, Guillaume Lample, Lucile Saulnier, Lélio~Renard Lavaud, Marie-Anne Lachaux, Pierre Stock, Teven~Le Scao, Thibaut Lavril, Thomas Wang, Timothée Lacroix, and William~El Sayed.
\newblock Mistral 7b, 2023.
\newblock URL \url{https://arxiv.org/abs/2310.06825}.

\bibitem[Lai et~al.(2024)Lai, Tian, Chen, Yang, Peng, and Jia]{lai2024stepdpo}
Xin Lai, Zhuotao Tian, Yukang Chen, Senqiao Yang, Xiangru Peng, and Jiaya Jia.
\newblock Step-dpo: Step-wise preference optimization for long-chain reasoning of llms.
\newblock \emph{arXiv preprint arXiv:2406.18629}, 2024.

\bibitem[Lewkowycz et~al.(2022)Lewkowycz, Andreassen, Dohan, Dyer, Michalewski, Ramasesh, Slone, Anil, Schlag, Gutman-Solo, et~al.]{lewkowycz2022solvingMinervaMath}
Aitor Lewkowycz, Anders Andreassen, David Dohan, Ethan Dyer, Henryk Michalewski, Vinay Ramasesh, Ambrose Slone, Cem Anil, Imanol Schlag, Theo Gutman-Solo, et~al.
\newblock Solving quantitative reasoning problems with language models.
\newblock In \emph{NeurIPS}, pp.\  3843--3857, 2022.

\bibitem[Light et~al.(2024)Light, Cai, Chen, Wang, Chen, Cheng, Yue, and Hu]{light2024strategist}
Jonathan Light, Min Cai, Weiqin Chen, Guanzhi Wang, Xiusi Chen, Wei Cheng, Yisong Yue, and Ziniu Hu.
\newblock Strategist: Learning strategic skills by llms via bi-level tree search.
\newblock \emph{arXiv preprint arXiv:2408.10635}, 2024.

\bibitem[Light et~al.(2025{\natexlab{a}})Light, Cheng, Yue, Oyamada, Wang, Paternain, and Chen]{light2025disc}
Jonathan Light, Wei Cheng, Wu~Yue, Masafumi Oyamada, Mengdi Wang, Santiago Paternain, and Haifeng Chen.
\newblock Disc: Dynamic decomposition improves llm inference scaling.
\newblock \emph{arXiv preprint arXiv:2502.16706}, 2025{\natexlab{a}}.

\bibitem[Light et~al.(2025{\natexlab{b}})Light, Wu, Sun, Yu, Zhao, Hu, Chen, Cheng, et~al.]{light2024scattered}
Jonathan Light, Yue Wu, Yiyou Sun, Wenchao Yu, Xujiang Zhao, Ziniu Hu, Haifeng Chen, Wei Cheng, et~al.
\newblock Scattered forest search: Smarter code space exploration with llms.
\newblock In \emph{ICLR}, 2025{\natexlab{b}}.

\bibitem[Lightman et~al.(2023)Lightman, Kosaraju, Burda, Edwards, Baker, Lee, Leike, Schulman, Sutskever, and Cobbe]{lightman2023let}
Hunter Lightman, Vineet Kosaraju, Yura Burda, Harri Edwards, Bowen Baker, Teddy Lee, Jan Leike, John Schulman, Ilya Sutskever, and Karl Cobbe.
\newblock Let's verify step by step.
\newblock \emph{arXiv preprint arXiv:2305.20050}, 2023.

\bibitem[Liu et~al.(2025{\natexlab{a}})Liu, Chen, Li, Qi, Pang, Du, Lee, and Lin]{liu2025understanding}
Zichen Liu, Changyu Chen, Wenjun Li, Penghui Qi, Tianyu Pang, Chao Du, Wee~Sun Lee, and Min Lin.
\newblock Understanding r1-zero-like training: A critical perspective.
\newblock \emph{arXiv preprint arXiv:2503.20783}, 2025{\natexlab{a}}.

\bibitem[Liu et~al.(2025{\natexlab{b}})Liu, Wang, Xu, Ma, Ruan, Li, Liu, and Wu]{liu2025dsgrm}
Zijun Liu, Peiyi Wang, Runxin Xu, Shirong Ma, Chong Ruan, Peng Li, Yang Liu, and Yu~Wu.
\newblock Inference-time scaling for generalist reward modeling.
\newblock \emph{arXiv preprint arXiv:2504.02495}, 2025{\natexlab{b}}.

\bibitem[Mnih et~al.(2013)Mnih, Kavukcuoglu, Silver, Graves, Antonoglou, Wierstra, and Riedmiller]{mnih2013playingDQN}
Volodymyr Mnih, Koray Kavukcuoglu, David Silver, Alex Graves, Ioannis Antonoglou, Daan Wierstra, and Martin Riedmiller.
\newblock Playing atari with deep reinforcement learning.
\newblock \emph{arXiv preprint arXiv:1312.5602}, 2013.

\bibitem[Mnih et~al.(2016)Mnih, Badia, Mirza, Graves, Lillicrap, Harley, Silver, and Kavukcuoglu]{mnih2016asynchronousA2C}
Volodymyr Mnih, Adria~Puigdomenech Badia, Mehdi Mirza, Alex Graves, Timothy Lillicrap, Tim Harley, David Silver, and Koray Kavukcuoglu.
\newblock Asynchronous methods for deep reinforcement learning.
\newblock In \emph{ICML}, pp.\  1928--1937, 2016.

\bibitem[Ouyang et~al.(2022)Ouyang, Wu, Jiang, Almeida, Wainwright, Mishkin, Zhang, Agarwal, Slama, Ray, et~al.]{ouyang2022training}
Long Ouyang, Jeffrey Wu, Xu~Jiang, Diogo Almeida, Carroll Wainwright, Pamela Mishkin, Chong Zhang, Sandhini Agarwal, Katarina Slama, Alex Ray, et~al.
\newblock Training language models to follow instructions with human feedback.
\newblock In \emph{NeurIPS}, pp.\  27730--27744, 2022.

\bibitem[Qin et~al.(2024)Qin, Li, Zou, Liu, Xia, Huang, Ye, Yuan, Liu, Li, et~al.]{qin2024o1}
Yiwei Qin, Xuefeng Li, Haoyang Zou, Yixiu Liu, Shijie Xia, Zhen Huang, Yixin Ye, Weizhe Yuan, Hector Liu, Yuanzhi Li, et~al.
\newblock O1 replication journey: A strategic progress report--part 1.
\newblock \emph{arXiv preprint arXiv:2410.18982}, 2024.

\bibitem[Rafailov et~al.(2024)Rafailov, Sharma, Mitchell, Manning, Ermon, and Finn]{rafailov2024direct}
Rafael Rafailov, Archit Sharma, Eric Mitchell, Christopher~D Manning, Stefano Ermon, and Chelsea Finn.
\newblock Direct preference optimization: Your language model is secretly a reward model.
\newblock In \emph{NeurIPS}, 2024.

\bibitem[Schulman et~al.(2017)Schulman, Wolski, Dhariwal, Radford, and Klimov]{schulman2017ppo}
John Schulman, Filip Wolski, Prafulla Dhariwal, Alec Radford, and Oleg Klimov.
\newblock Proximal policy optimization algorithms.
\newblock \emph{arXiv preprint arXiv:1707.06347}, 2017.

\bibitem[Shao et~al.(2024)Shao, Wang, Zhu, Xu, Song, Bi, Zhang, Zhang, Li, Wu, et~al.]{shao2024deepseekmath}
Zhihong Shao, Peiyi Wang, Qihao Zhu, Runxin Xu, Junxiao Song, Xiao Bi, Haowei Zhang, Mingchuan Zhang, YK~Li, Y~Wu, et~al.
\newblock Deepseekmath: Pushing the limits of mathematical reasoning in open language models.
\newblock \emph{arXiv preprint arXiv:2402.03300}, 2024.

\bibitem[Srivastava \& Aggarwal(2025)Srivastava and Aggarwal]{srivastava2025technical}
Saksham~Sahai Srivastava and Vaneet Aggarwal.
\newblock A technical survey of reinforcement learning techniques for large language models.
\newblock \emph{arXiv preprint arXiv:2507.04136}, 2025.

\bibitem[Sutton(1988)]{Sutton1988}
Richard~S. Sutton.
\newblock Learning to predict by the methods of temporal differences.
\newblock \emph{Machine Learning}, 3\penalty0 (1):\penalty0 9--44, 1988.
\newblock \doi{10.1007/BF00115009}.

\bibitem[Team et~al.(2023)Team, Anil, Borgeaud, Wu, Alayrac, Yu, Soricut, Schalkwyk, Dai, Hauth, et~al.]{team2023gemini}
Gemini Team, Rohan Anil, Sebastian Borgeaud, Yonghui Wu, Jean-Baptiste Alayrac, Jiahui Yu, Radu Soricut, Johan Schalkwyk, Andrew~M Dai, Anja Hauth, et~al.
\newblock Gemini: a family of highly capable multimodal models.
\newblock \emph{arXiv preprint arXiv:2312.11805}, 2023.

\bibitem[Tesauro(1991)]{tesauro1991practical}
Gerald Tesauro.
\newblock Practical issues in temporal difference learning.
\newblock In \emph{NeurIPS}, 1991.

\bibitem[Tesauro(2002)]{tesauro2002programming}
Gerald Tesauro.
\newblock Programming backgammon using self-teaching neural nets.
\newblock \emph{Artificial Intelligence}, 134\penalty0 (1-2):\penalty0 181--199, 2002.

\bibitem[Tie et~al.(2025)Tie, Zhao, Song, Wei, Zhou, Dai, Yin, Yang, Yan, Su, et~al.]{tie2025large}
Guiyao Tie, Zeli Zhao, Dingjie Song, Fuyang Wei, Rong Zhou, Yurou Dai, Wen Yin, Zhejian Yang, Jiangyue Yan, Yao Su, et~al.
\newblock Large language models post-training: Surveying techniques from alignment to reasoning.
\newblock \emph{arXiv preprint arXiv:2503.06072}, 2025.

\bibitem[Uesato et~al.(2022)Uesato, Kushman, Kumar, Song, Siegel, Wang, Creswell, Irving, and Higgins]{uesato2022solving}
Jonathan Uesato, Nate Kushman, Ramana Kumar, Francis Song, Noah Siegel, Lisa Wang, Antonia Creswell, Geoffrey Irving, and Irina Higgins.
\newblock Solving math word problems with process-and outcome-based feedback.
\newblock \emph{arXiv preprint arXiv:2211.14275}, 2022.

\bibitem[Wang et~al.(2024)Wang, Li, Shao, Xu, Dai, Li, Chen, Wu, and Sui]{wang2023math}
Peiyi Wang, Lei Li, Zhihong Shao, RX~Xu, Damai Dai, Yifei Li, Deli Chen, Y~Wu, and Zhifang Sui.
\newblock Math-shepherd: A label-free step-by-step verifier for llms in mathematical reasoning.
\newblock In \emph{ACL}, pp.\  9426–9439, 2024.

\bibitem[Wei et~al.(2022)Wei, Wang, Schuurmans, Bosma, Xia, Chi, Le, Zhou, et~al.]{wei2022chain}
Jason Wei, Xuezhi Wang, Dale Schuurmans, Maarten Bosma, Fei Xia, Ed~Chi, Quoc~V Le, Denny Zhou, et~al.
\newblock Chain-of-thought prompting elicits reasoning in large language models.
\newblock In \emph{NeurIPS}, pp.\  24824--24837, 2022.

\bibitem[Xia et~al.(2024)Xia, Zhang, Liao, Hou, Sun, Li, Fu, and Dong]{xia2024scenegenagent}
Xiao Xia, Dan Zhang, Zibo Liao, Zhenyu Hou, Tianrui Sun, Jing Li, Ling Fu, and Yuxiao Dong.
\newblock Scenegenagent: Precise industrial scene generation with coding agent.
\newblock \emph{arXiv preprint arXiv:2410.21909}, 2024.

\bibitem[Yang et~al.(2024{\natexlab{a}})Yang, Yang, Zhang, Hui, Zheng, Yu, Li, Liu, Huang, Wei, et~al.]{yang2024qwen2}
An~Yang, Baosong Yang, Beichen Zhang, Binyuan Hui, Bo~Zheng, Bowen Yu, Chengyuan Li, Dayiheng Liu, Fei Huang, Haoran Wei, et~al.
\newblock Qwen2.5 technical report.
\newblock \emph{arXiv preprint arXiv:2412.15115}, 2024{\natexlab{a}}.

\bibitem[Yang et~al.(2024{\natexlab{b}})Yang, Zhang, Hui, Gao, Yu, Li, Liu, Tu, Zhou, Lin, et~al.]{yang2024qwen2math}
An~Yang, Beichen Zhang, Binyuan Hui, Bofei Gao, Bowen Yu, Chengpeng Li, Dayiheng Liu, Jianhong Tu, Jingren Zhou, Junyang Lin, et~al.
\newblock Qwen2.5-math technical report: Toward mathematical expert model via self-improvement.
\newblock \emph{arXiv preprint arXiv:2409.12122}, 2024{\natexlab{b}}.

\bibitem[Yao et~al.(2024)Yao, Yu, Zhao, Shafran, Griffiths, Cao, and Narasimhan]{yao2024tree}
Shunyu Yao, Dian Yu, Jeffrey Zhao, Izhak Shafran, Tom Griffiths, Yuan Cao, and Karthik Narasimhan.
\newblock Tree of thoughts: Deliberate problem solving with large language models.
\newblock In \emph{NeurIPS}, 2024.

\bibitem[Yeo et~al.(2025)Yeo, Tong, Niu, Neubig, and Yue]{yeo2025demystifying}
Edward Yeo, Yuxuan Tong, Morry Niu, Graham Neubig, and Xiang Yue.
\newblock Demystifying long chain-of-thought reasoning in llms.
\newblock In \emph{ICML}, 2025.

\bibitem[Yu et~al.(2024)Yu, Jiang, Shi, Yu, Liu, Zhang, Kwok, Li, Weller, and Liu]{yu2023metamath}
Longhui Yu, Weisen Jiang, Han Shi, Jincheng Yu, Zhengying Liu, Yu~Zhang, James~T Kwok, Zhenguo Li, Adrian Weller, and Weiyang Liu.
\newblock Metamath: Bootstrap your own mathematical questions for large language models.
\newblock In \emph{ICLR}, 2024.

\bibitem[Zeng et~al.(2025)Zeng, Huang, Liu, Liu, He, Ma, and He]{zeng2025simplerl}
Weihao Zeng, Yuzhen Huang, Qian Liu, Wei Liu, Keqing He, Zejun Ma, and Junxian He.
\newblock Simplerl-zoo: Investigating and taming zero reinforcement learning for open base models in the wild.
\newblock \emph{arXiv preprint arXiv:2503.18892}, 2025.

\bibitem[Zhang et~al.(2024{\natexlab{a}})Zhang, Hu, Zhoubian, Du, Yang, Wang, Yue, Dong, and Tang]{sciglm}
Dan Zhang, Ziniu Hu, Sining Zhoubian, Zhengxiao Du, Kaiyu Yang, Zihan Wang, Yisong Yue, Yuxiao Dong, and Jie Tang.
\newblock Sciinstruct: a self-reflective instruction annotated dataset for training scientific language models.
\newblock In \emph{NeurIPS}, pp.\  1443--1473, 2024{\natexlab{a}}.

\bibitem[Zhang et~al.(2024{\natexlab{b}})Zhang, Zhoubian, Hu, Yue, Dong, and Tang]{zhang2024rest}
Dan Zhang, Sining Zhoubian, Ziniu Hu, Yisong Yue, Yuxiao Dong, and Jie Tang.
\newblock Rest-mcts*: Llm self-training via process reward guided tree search.
\newblock In \emph{NeurIPS}, pp.\  64735--64772, 2024{\natexlab{b}}.

\bibitem[Zhang et~al.(2025)Zhang, Feng, Xue, Wang, Dong, and Tang]{zhang2025parameter}
Dan Zhang, Tao Feng, Lilong Xue, Yuandong Wang, Yuxiao Dong, and Jie Tang.
\newblock Parameter-efficient fine-tuning for foundation models.
\newblock \emph{arXiv preprint arXiv:2501.13787}, 2025.

\bibitem[Zheng et~al.(2023)Zheng, Xia, Zou, Dong, Wang, Xue, Shen, Wang, Wang, Li, Su, Yang, and Tang]{codegeex}
Qinkai Zheng, Xiao Xia, Xu~Zou, Yuxiao Dong, Shan Wang, Yufei Xue, Lei Shen, Zihan Wang, Andi Wang, Yang Li, Teng Su, Zhilin Yang, and Jie Tang.
\newblock Codegeex: A pre-trained model for code generation with multilingual benchmarking on humaneval-x.
\newblock In \emph{SIGKDD}, pp.\  5673–5684, 2023.

\end{thebibliography}
\bibliographystyle{iclr2026_conference}

\clearpage
\appendix
\section{Statement of LLM Usage}
This manuscript was prepared by the authors, who take full responsibility for its content. Large language models (ChatGPT, etc.) were used solely for language polishing and grammar suggestions. No generated text or analysis was included without human verification.

\section{Table of Notations}
% \usepackage{booktabs}
\begin{table}[ht!]
\caption{Table of Notations}
\label{tab:notations}
\begin{tabular}{@{}ll@{}}
\toprule
Symbol                       & Meaning     \\ \midrule
$N$  & number of generations for Best-of-$N$\\
$\mathcal{S}$ & state space \\
$\mathcal{A}$ & action space\\
$f$ & transition function \\
$R$ & reward function ($R$: $\mathcal{S} \times \mathcal{A} \rightarrow r, r \in \mathbb{R}$)\\
$s_t$ & state (token sequence or context for LLM) at step t\\
$a_t$ & action (newly generated token) at step t \\
$(x_0, \dots, x_L)$          & input prompt\\
$(y_0, \dots, y_{t-1})$    & sequence of generation tokens \\
$T$ & terminal step \\
$\rho_{\pi}$ & trajectory distribution \\
$\beta$& KL coefficient \\
$q$ & query \\
$O = o_1 \dots o_G$ & response \\
$r_t$ & reward at step $t$ \\
$n$ & $n$-step TD \\
$\gamma$ & discount factor\\
$V$ & value function \\
$\alpha$ & step size in TD \\
% $C$ & cosine reward function \\
$v_t$ & TD target   \\
$\tilde{v}_t$ & clamped TD target   \\
$V_t$ & TD target at terminal step \\
$\hat{g}$& predicted answer \\
$g$& ground truth answer                  \\
$R_\text{verifiable}$ & verifiable reward function \\
$R_\text{PRM}$ & PRM reward function \\
$\text{PRM}_{\phi}$ & PRM, $\phi$ refers to the parameter \\
$a$ & coefficient for ``TDRL'' \\
$\hat{A}_{i,j}$ & advantage for the $j$-th token of each $\tau_i$ using group relative advantage estimation \\
$\tau_i$ & in Alg.~\ref{alg:grpo-tdrm}, the $i$-th trajectory in batch $\mathcal{D}_b$ sampled from task prompts $\mathcal{D}_\text{policy}$ \\
$\tau_{\textrm{PRM}}$ & trajectory in TD PRM dataset $\mathcal{D}_\text{PRM}$ \\
$U$ & discounted return \\
\bottomrule
\end{tabular}
\end{table}

\section{Related Work}
\subsection{Reasoning Process Reward}
LLMs have achieved significant performance improvement in advanced complex reasoning scenarios~\cite{jaech2024openai, qin2024o1, light2024strategist} through step-by-step reasoning.
For example, CoT~\cite{wei2022chain}, ToT~\cite{yao2024tree}, SFS~\cite{light2024scattered}, and MCTS~\cite{zhang2024rest} have progressed in reasoning tasks by analyzing complex questions and providing guidance for models to obtain correct solutions.
Uesato~\cite{uesato2022solving} and Light et al.~\cite{lightman2023let} propose the ORM that detects the final result and PRM that provides the feedback for intermediate reasoning steps, and demonstrate that PRM is more effective than ORM for obtaining correct step-level process and avoiding the false positive steps that match with final correct answer with incorrect solutions.
Step-DPO~\cite{lai2024stepdpo} checks step-by-step answers and collects positive and negative step-level solutions for training direct preference optimization~\cite{rafailov2024direct} rather than evaluating the correctness of whole solutions and final answer.
Math-shepherd~\cite{wang2023math} and ReST-MCTS*~\cite{zhang2024rest} introduce reinforced training by integrating process reward with tree search to collect high-quality reasoning paths for LLMs in mathematical reasoning.
Despite their effectiveness, the research to obtain automated, more correct, and label-free process rewards remains unexplored.
To implement this goal, we propose a new reward function for process reward optimization and online training of LLMs.

\subsection{Reinforcement Learning Training of LLMs}
Direct preference optimization (DPO)~\cite{rafailov2024direct} optimizes models by learning positive and negative pairs.
Compared to DPO, Proximal Policy Optimization (PPO)~\cite{schulman2017ppo} is an effective online RLHF algorithm but requires high GPU memory and is challenged in real-use scenarios.
To fill the gap, reinforce leave-one-out (RLOO)~\cite{ahmadian2024rloo} is proposed to load the policy, reference, and reward models to memory, and model the entire completed token as a single action.

\subsection{Temporal Difference in Reinforcement Learning}
TD plays a vital role in connecting model-based and model-free methods within RL, estimating state values by merging immediate rewards with discounted future state values. 
The foundational 1-step TD algorithm~\cite{Sutton1988} updates state value estimates using TD errors, enabling agents to learn optimal policies online. 
TD methods have also been integrated with policy search techniques, resulting in TD-based policy gradient algorithms such as A2C~\cite{mnih2016asynchronousA2C} that leverage TD errors to optimize policies, achieving great success in game playing~\cite{tesauro1991practical, tesauro2002programming}.
In the deep RL realm, DQN~\cite{mnih2013playingDQN} and its variants utilize TD learning to train neural networks approximating the Q-function.
Therefore, TD learning is a natural method for training reliable and smoother reward models for RL training.

\section{Algorithm Details}
\begin{algorithm}[h!]
\caption{Backward view of TD-$\lambda$ for PRM training}
\label{alg:td_lambda}
\textbf{Notation:} $s_t$: state; $a_t$: action; $r_t$: reward; $V(s_t)$: state value; $\hat{V}(s_t)$: updated value estimate; $e(s)$: eligibility trace; $\delta$: TD error; $\pi$: policy \\
\textbf{Input:} Dataset $\mathcal{D}_{\textrm{PRM}}$ of trajectories with rewards $\{r_t\}_{t=1}^{T}$; process reward model PRM$_\phi$ with parameters $\phi$; discount factor $\gamma$; step size $n$; eligibility trace decay rate $\lambda$
\begin{algorithmic}[1]
    \STATE Initialize total loss: $\mathcal{L} \leftarrow 0$
    \FOR{each trajectory $\tau = \{(s_1, r_1), \dots, (s_T, r_T)\}$ in $\mathcal{D}_{\textrm{PRM}}$}
    \STATE Initialize value estimates: $\hat{V}(s_t) \leftarrow \text{PRM}_\phi(s_{t})$, $\forall s_t$
    \STATE Initialize eligibility traces: $e(s_t) \leftarrow 0$, $\forall s_t$
    
    \FOR{$t = 1$ to $T-1$}
        \STATE $V(s_{t+1}) \leftarrow \text{PRM}_\phi(s_{t+1})$
        \STATE $e(s_t) \leftarrow \gamma \lambda \cdot e(s_t)$
        \STATE $e(s_t) \leftarrow e(s_t) + 1$
        \STATE $\delta \leftarrow r_t + \gamma \cdot \hat{V}(s_{t+1}) - \hat{V}(s_t)$
        \FOR{$j=1$ to $t$}
            \STATE $\hat{V}(s_j) \leftarrow \hat{V}(s_j) + \delta \cdot e(s_t)$
        \ENDFOR
    \ENDFOR
   \STATE $\mathcal{L}\leftarrow\sum_{t=1}^{T-1} 
\text{CE}\left(\sigma(V(s_t)), \underbrace{\sigma(\hat{V}(s_t))}_{\text{TD target}}\right) + \text{CE}\left(\sigma(V(s_T)), \underbrace{r_T}_{\text{TD target at terminal state}}\right)$
    \STATE $\mathcal{L}.\texttt{backward()}$
    \STATE $\mathcal{L}\leftarrow0$
    \ENDFOR
    
\end{algorithmic}
\end{algorithm}
\begin{algorithm}[ht!]
    \small
    \caption{$n$-step TD for PRM training}
    \label{alg:tdn}
    \textbf{Notation:} $s_t$: the $t$-th reasoning step; $r_t$: reward at $s_t$; $V(s_t)$: state value at $s_t$; $\sigma$: sigmoid; CE: cross-entropy loss; \texttt{clamp}: clamps value in [0, 1]; $n$: TD step size \\
    \textbf{Input:} Dataset $\mathcal{D}_{\textrm{PRM}}$ of trajectories with rewards $\{r_t\}_{t=1}^{T}$; process reward model PRM$_\phi$ with parameters $\phi$; discount factor $\gamma$; step size $n$; $U$: discounted return over a set of steps
    \begin{algorithmic}[1]
        \STATE Initialize total loss: $\mathcal{L}_{\text{total}} \leftarrow 0$
        \FOR{each trajectory $\tau = \{(s_1, r_1), \dots, (s_T, r_T)\}$ in $\mathcal{D}_{\textrm{PRM}}$}
            \FOR{$t = 1$ to $T$}
                \STATE $V(s_t) \leftarrow \sigma(\text{PRM}_\phi(s_t))$
                \STATE $U \leftarrow 0$
                \FOR{$k = 0$ to $n-1$}
                    \IF{$t+k \leq T$}
                        \STATE $U \leftarrow U + \gamma^k \cdot r_{t+k}$
                    \ENDIF
                \ENDFOR
                \IF{$t+n \leq T$}
                    \STATE $V(s_{t+n}) \leftarrow \sigma(\text{PRM}_\phi(s_{t+n}))$
                    \STATE $U \leftarrow U + \gamma^n \cdot V(s_{t+n})$
                \ENDIF
                \STATE $\mathcal{L}_t \leftarrow \text{CE}\left(V(s_t), \underbrace{\texttt{clamp}(U, 0, 1)}_{\text{TD target}}\right)$
                \STATE $\mathcal{L}_{\text{total}} \leftarrow \mathcal{L}_{\text{total}} + \mathcal{L}_t$
            \ENDFOR
            \STATE $\mathcal{L}_{\text{total}}.\texttt{backward()}$
            \STATE $\mathcal{L}_{\text{total}}\leftarrow0$
        \ENDFOR
    \end{algorithmic}
\end{algorithm}

\section{Comparison of reward models}
\begin{figure}[h!]
    \centering
    \includegraphics[width=0.95\linewidth]{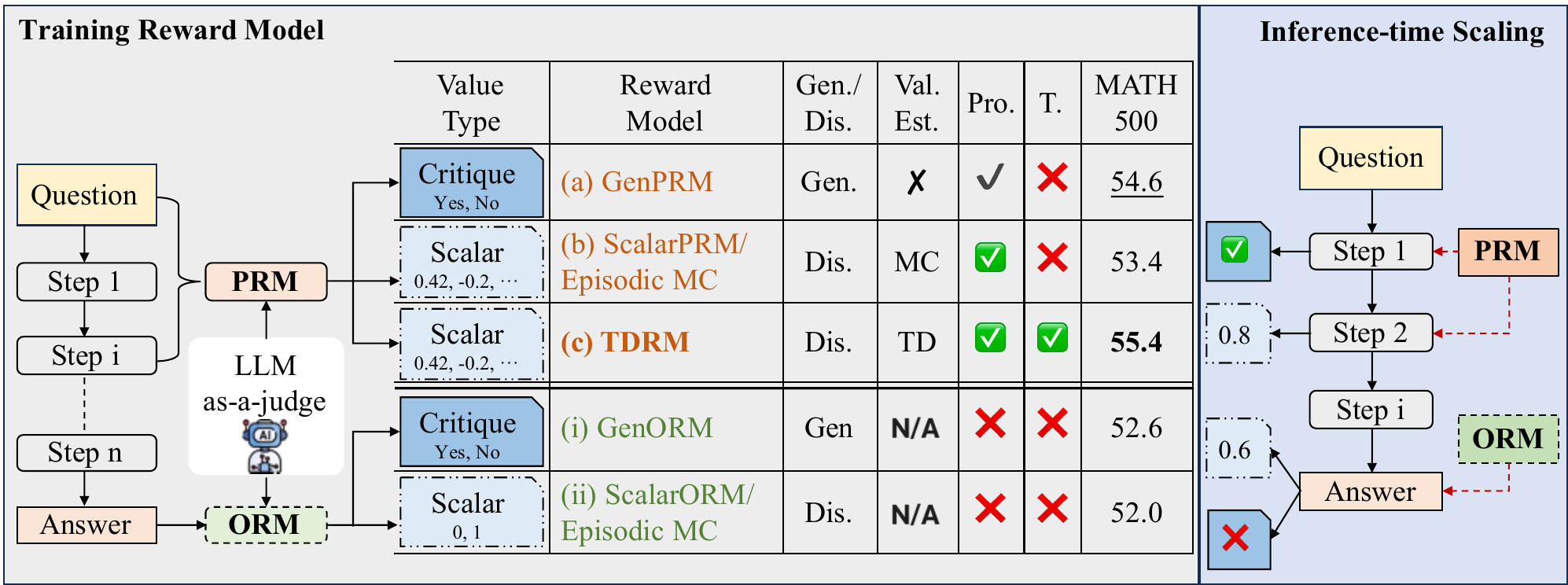}
    \vspace{-0.2cm}
    \caption{Comparison of recent reward models and \model. Gen. and Dis. denote Generative and Discriminative. 
    Val. Est. denotes the method of value estimation. 
    Pro. and T. denote process and temporality. 
    MC and TD denote Monte Carlo and temporal difference.
    }
    \label{fig: RMs}
\end{figure}

Figure~\ref{fig: RMs} provides a comprehensive comparison of recent reward models from various perspectives (e.g., value type, reward model, value estimation, temporal consistency, and MATH-500 results).

\section{Experiment Details}
\subsection{Experiment Settings}
\label{appendix: experiment}

\vpara{Evaluation Metrics}
Best-of-$N$ Sampling is designed to balance diversity and optimality across outputs. 
In our experiments, $N$ is set to $\{128, 1024\}$.
While Greedy Search~\cite{light2025disc} is efficient, it risks suboptimal results due to locally optimal decisions. 
For a fair comparison and a more thorough study, we pre-generate reasoning trajectories using \texttt{Mistral-7B-Instruct-v0.2} for GSM8K (128 outputs for questions). 
For MATH-500, we use \texttt{RLHFlow/Mistral-MATH500-Test}~\cite{dong2024rlhflow} from \texttt{hugging-face}, which contains 1,024 outputs for each question, generated with a \texttt{Mistral-7B} model~\cite{jiang2023mistral7b}, which has been fine-tuned on \texttt{MetaMath}~\cite{yu2023metamath}.
For Greedy Search, we set the sampling temperature and backbone as $0.4$ and Qwen2.5-Math-7B for generation, and run all the experiments three times to mitigate randomness.

\vpara{Dataset for \model Training.}
For TD-based PRM training, we use the \texttt{RLHFlow/Mistral-PRM-Data}, which contains step-by-step reasoning trajectories with corresponding correctness labels for intermediate steps. 
For online RL training, we utilized MATH Level-3 data~\cite{hendrycks2021measuring}, which comprises 2,500 problem prompts designed to evaluate advanced mathematical reasoning capabilities.

\vpara{Baselines.}
In verification experiments, we train our baseline using the same Cross-Entropy loss, whereas the target is instead the hard label $Y$.
ScalarORM refers to training with only the terminal state, and ScalarPRM incorporates both the intermediate and terminal states. 
For this setting, we train a ScalarORM using \texttt{RLHFlow/Mistral-ORM-Data} and a ScalarPRM using \texttt{RLHFlow/Mistral-PRM-Data}. 
For RL training comparisons, we benchmark \model with SimpleRL-zoo~\cite{zeng2025simplerl}, GRPO in DeepSeek-Math~\cite{shao2024deepseekmath}, Dr.GRPO~\cite{liu2025understanding}, and OpenReasoner-Zero~\cite{hu2025open}.

\vpara{RL Training Setting.}
We conduct online RL training across 4 series of models, including Qwen2.5-(0.5B, 1.5B)~\cite{yang2024qwen2}, GLM4-9B-0414, GLM-Z1-9B-0414~\cite{glm2024chatglm}, Qwen2.5-Math-(1.5B, 7B)~\cite{yang2024qwen2math}, and DS-R1-Distill-Qwen-(1.5B, 7B)~\cite{deepseekai2025deepseekr1}. 
In the \model framework, the coefficient of $R_{\textrm{PRM}}$ is set to $a=0.2$. 
We run training with a total batch size of 56, divided equally across 7 GPUs, yielding 8 samples per GPU, to optimize compute and efficiency. 
The rest is used for online sampling with the number of rollouts set to 7, max completion length of 2048, and 1 epoch to mitigate overfitting risks.
The RL training framework of \model is developed from \texttt{huggingface/trl}. 

In our policy training experiments, aside from the same number of responses for each prompt, $7$, to estimate group relative advantage, we allocate different compute resources according to the model size, which leads to a choice of different batch sizes. For 7B and 3B models, we use 8 GPUs, 1 for sampling and 7 for training. 
The global batch size is $7~(\textrm{number of devices for gradient calculation})\times8~(\textrm{per device batch size})=56$. 
For 1.5B models, we use 4 GPUs, and the global batch size is $3\times14=42$. 
For 0.5B models, we use 2 GPUs, and the global batch size is $28\times1\times2~(\textrm{gradient accumulation steps})=56$. 
For \model training, we use a global batch size of $8~(\textrm{number of devices})\times16~(\textrm{per device batch size})\times2~(\textrm{gradient accumulation steps})=256$. As for GLM series models, we use \texttt{slime}\footnote{https://github.com/THUDM/slime} for training, with a global batch size of $256$, and $8$ responses for each prompt to estimate group relative advantage.

\begin{figure*}[t!]
    \centering
    % \begin{subfigure}[b]{0.46\textwidth}
    %     \centering
    %     \label{fig: td-search-qwen}
    %     \includegraphics[height=2.2in]{figs/tree_search_qwen.pdf}
    %     % \caption{}
    % \end{subfigure}
    % ~
    \begin{subfigure}[b]{0.46\textwidth}
        \centering
        \label{fig: td-search-mistral}
        \includegraphics[height=2.2in]{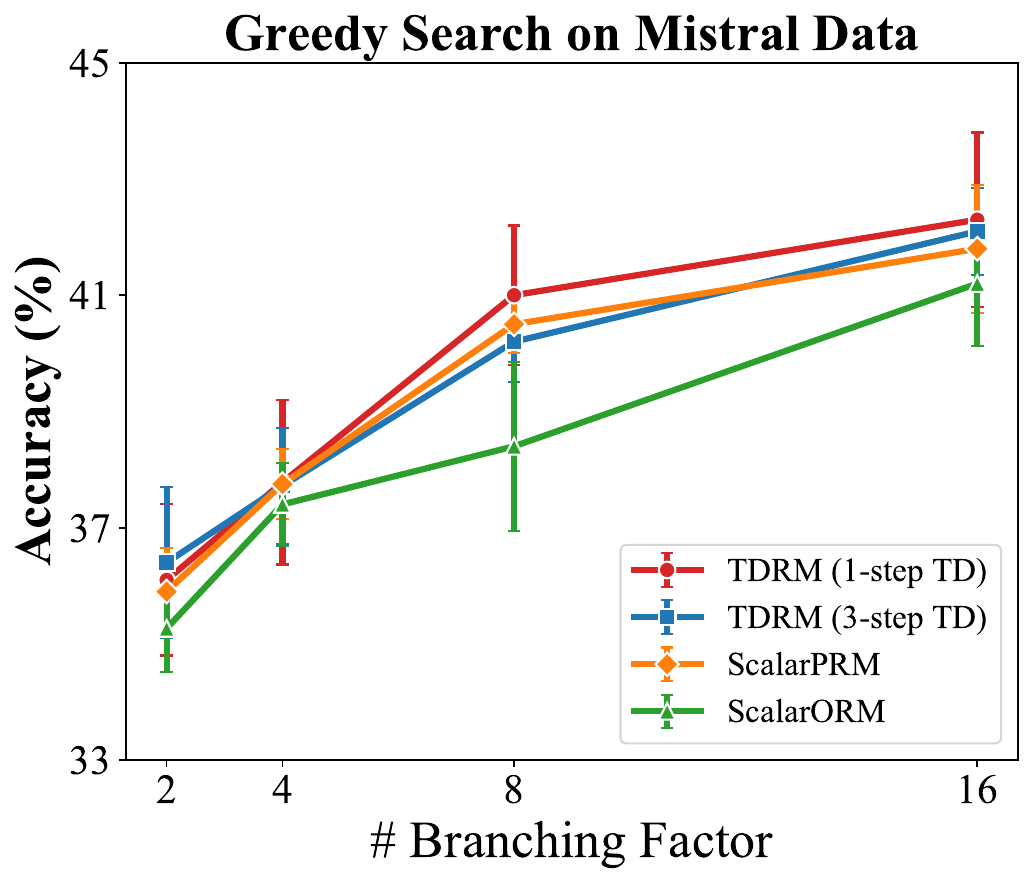}
        % \caption{}
    \end{subfigure}
    \vspace{-0.4cm}
    \caption{Results of greedy search on our PRM with TD.}
    \label{fig:td_search_mistral}
\end{figure*}

% \subsection{Reward Modeling Results}
% In tree search evaluations, as illustrated in Figure~\ref{fig:td_search}, \model again demonstrates superior performance and provides more accurate verification of reasoning trajectories.
% Moreover, \model exhibits enhanced reliability, with its accuracy improving as the number of search branches increases, indicating its effectiveness in navigating complex decision spaces. 

\begin{figure}[t!]
    \centering
    % \vspace{30pt}
    \includegraphics[height=1.6in]{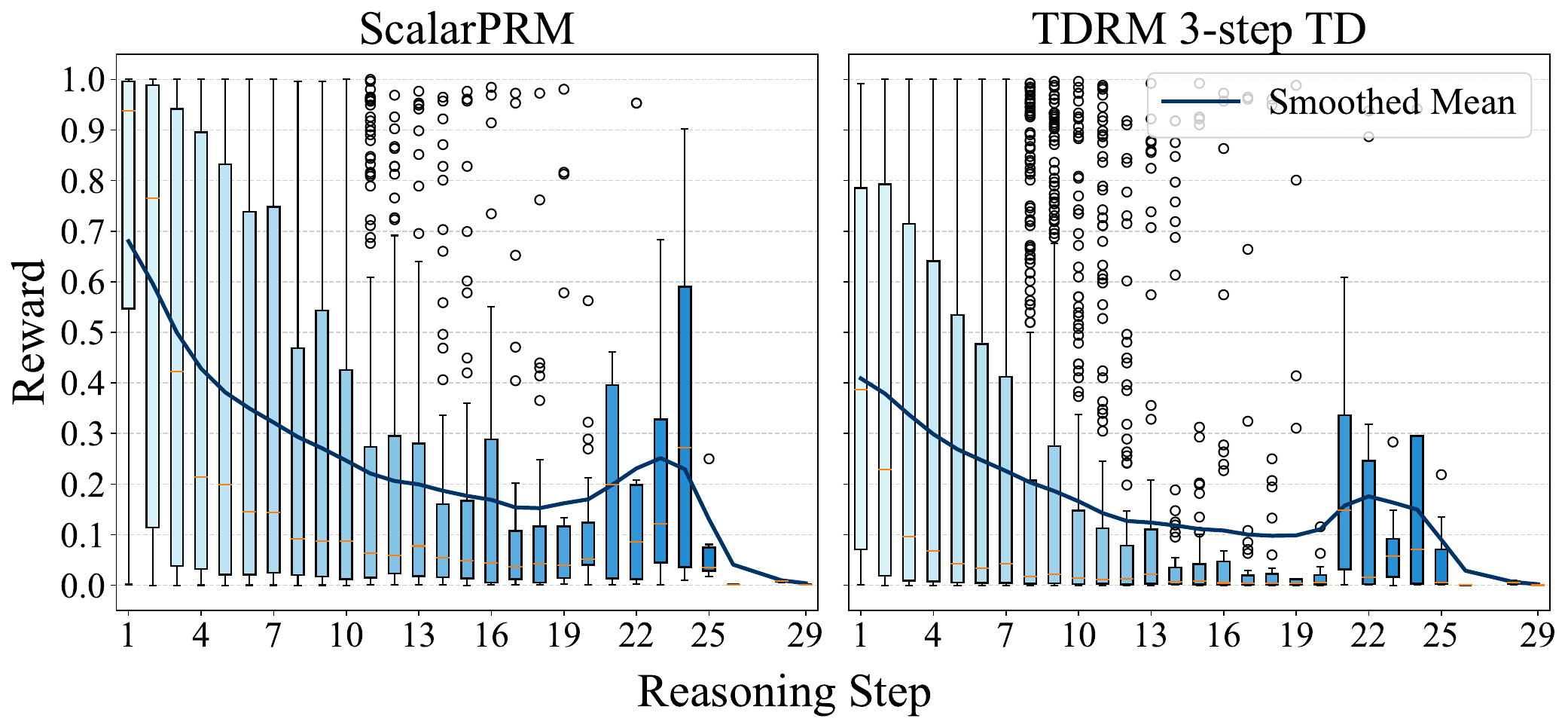}
    \hfill
    \includegraphics[height=1.6in]{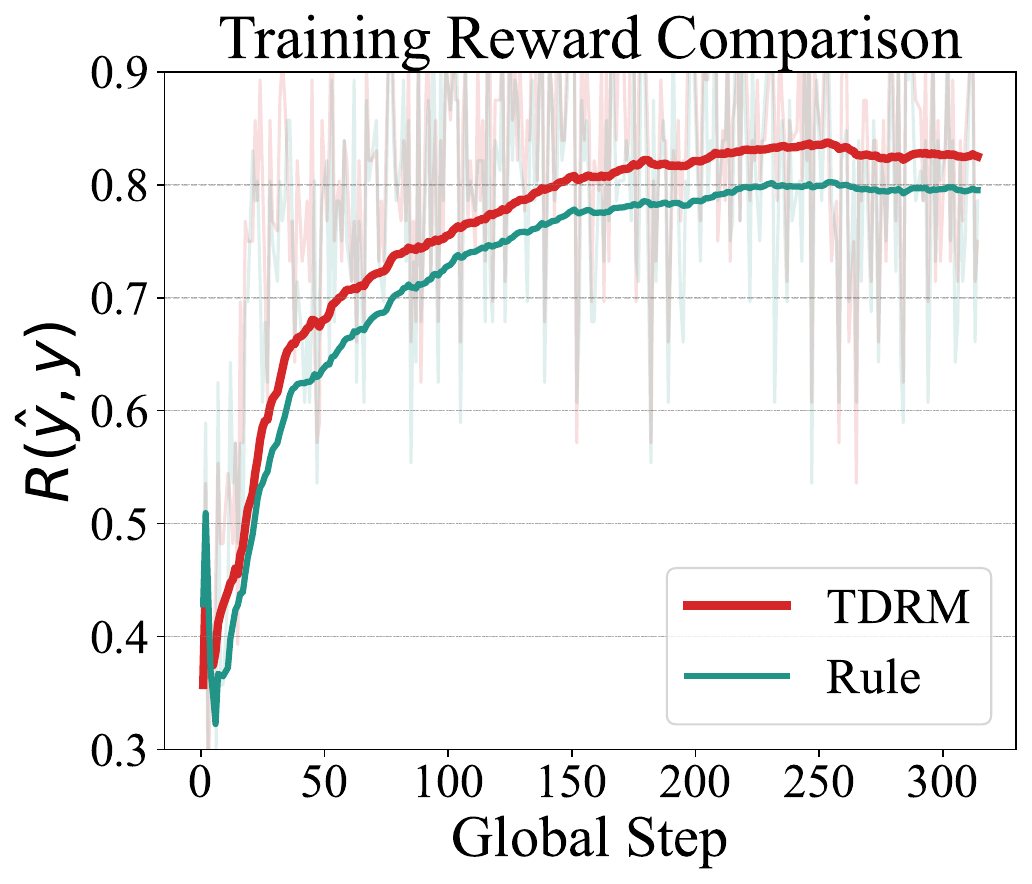}
    \vspace{-0.2cm}
    \caption{\textbf{Left}: Reward distribution over different reasoning steps. TDRM produces more stable and consistent reward estimates, reducing noisy spikes. \textbf{Right}: Dynamics of training reward. \model consistently yields higher rewards compared to the rule-based baseline, starting from the early steps of training.
    % \textcolor{red}{(YY: what is the conclusion of comparing the left two plots?)}
    }
    \label{fig:reward_distribution}
    % \vspace{-10pt}
\end{figure}

\subsection{Underlying Reward Distribution}
% \vpara{Reward Distribution.}
To better understand what a good reward model is like, we visualize the reward distribution over different reasoning steps in Figure~\ref{fig:reward_distribution}. 
This is similar to the smoothness analysis, while it is primarily focused on the distribution of state values, i.e., rewards of \model. 
As shown in Figure~\ref{fig:reward_distribution}, the trend of reward distribution is similar for both RMs, where it is in a ``U'' shape as the reasoning step increases, and then drops drastically as the reasoning step becomes much larger. This may reflect the underlying distribution of the dataset that we use to study.
% (see Appendix \ref{appendix:reward_distribution}).
However, the distribution of \model is smoother and more flat than ScalarPRM, indicating that it is more robust to the number of reasoning steps.

\begin{figure*}[t!]
    \centering
    \includegraphics[width=1.0\linewidth]{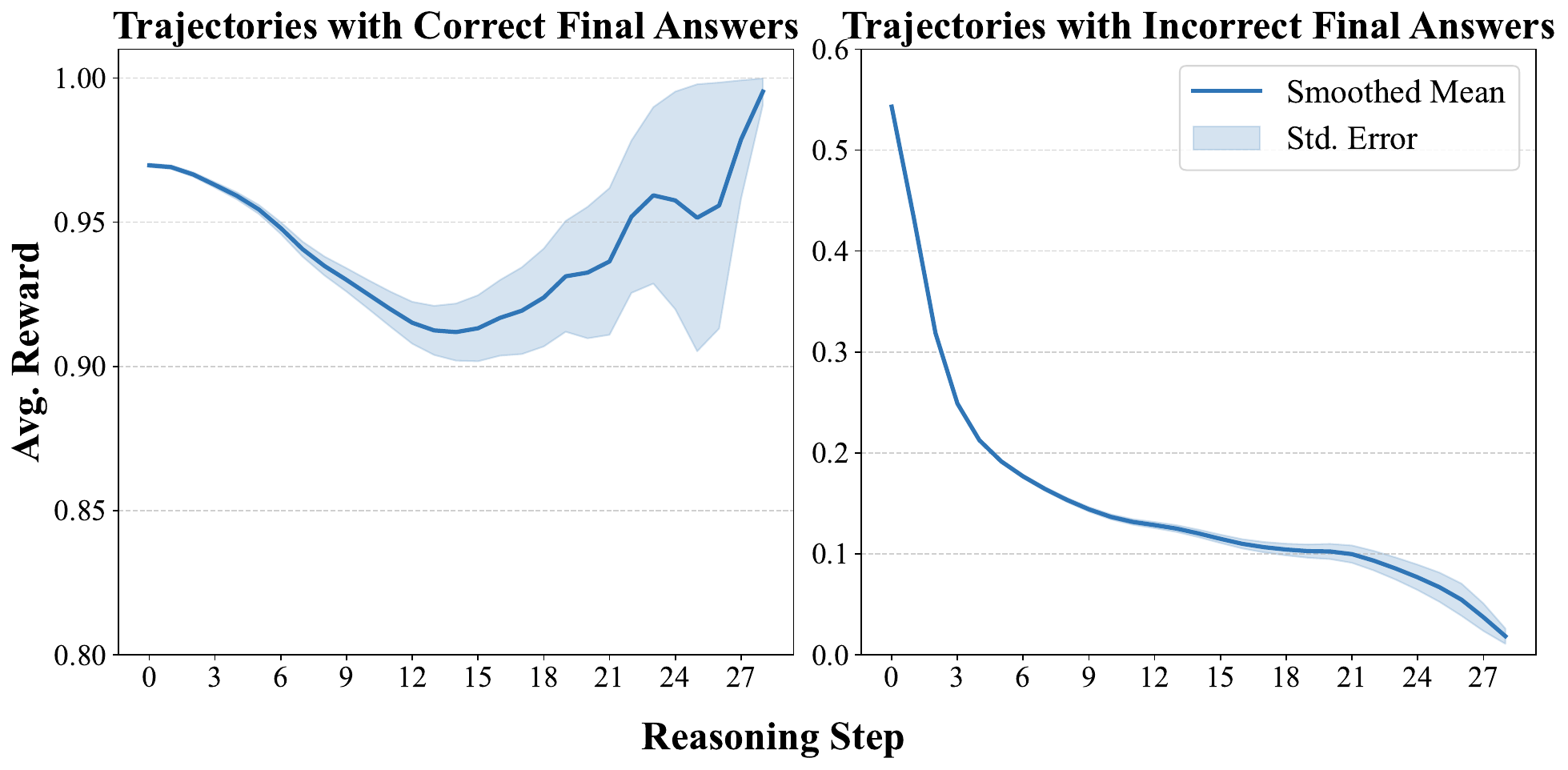}
    \caption{Training reward distribution. This figure shows the smoothed mean reward across different reasoning steps, using trajectories of correct and incorrect final answers separately.}
    % \vspace{-0.2cm}
    \label{fig: training_reward_distribution}
\end{figure*}

% \subsection{Underlying Reward Distribution}
% \label{appendix:reward_distribution}
% In Figure~\ref{fig:reward_distribution}, we compare the reward distribution of \model versus ScalarPRM, which shows a similar trend. 
% Here we provide additional details, including the reward distribution of the training data, and separately illustrate the reward distribution with the correct final answer versus those with incorrect answers.

% \vpara{Training Reward Distribution.}
% In Figure~\ref{fig:reward_distribution}, we saw that reward distributions of both ScalarPRM and \model first exhibit a ``U'' shape and then drop drastically as the number of reasoning steps gets large. 
To better understand the underlying mechanism, we first decompose the reward distribution in Figure~\ref{fig:reward_distribution} into two distributions, i.e., a distribution for trajectories with correct final answers and a distribution for trajectories with incorrect final answers. 
As shown in Figure~\ref{fig: training_reward_distribution}, the reward distribution of trajectories with correct answers exhibits a ``U'' shape, while that of trajectories with incorrect answers decreases as the number of reasoning steps increases.

\subsection{Template used in RL Training}
\definecolor{custom_blue}{RGB}{49,117,181}
\begin{tcolorbox}[colframe=custom_blue, colback=white, title=Prompt for implementation]
\texttt{<System>}\\
Please reason step by step, and put your final answer within \textbackslash boxed\{\}.\\
\texttt{</System>}\\
\texttt{<User>}\\
Question:\\
\textit{Input Question}\\
\texttt{<Assistant>}\\
Answer:\\
Let's think step by step.
\end{tcolorbox}

\subsection{Details of Verifiable Reward}
\label{sec: verifiable_reward}
Here we provide the concrete definition of \texttt{is\_equivalent} and \texttt{has\_boxed} of $R_\text{verifiable}$.
\begin{itemize}[leftmargin=*,itemsep=0pt,parsep=0.5em,topsep=0.3em,partopsep=0.3em]
    \item \texttt{is\_equivalent}: We only consider \boxed{\textrm{boxed answers}} wrapped within a \textbackslash boxed\{\}. And we calculate equivalence after normalizing both $\hat{g}$ and $g$, using a third-party package $\texttt{mathruler}$\footnote{https://github.com/hiyouga/MathRuler/tree/main}. If the answers are equivalent then return \texttt{True}, otherwise \texttt{False}.
    \item \texttt{has\_boxed}: To determine if the response has boxed answers and to extract the boxed answers, we use regex \texttt{``.*\textbackslash \textbackslash boxed\{.*\}.*''}. If the response matches the regex, then return \texttt{True}, otherwise \texttt{False}.
\end{itemize}

\subsection{Local Lipschitz Constant for Smoothness}
\label{app: Lipschitz_constant}
Inspired by the local Lipschitz Constant, we use the following formula to calculate the smoothness of PRMs:
\begin{equation}
L_{\text{smoothness}} = \frac{1}{|\mathcal{D}|} \sum_{(s_t, s_{t+1}) \in \mathcal{D}} \frac{|V(s_{t+1}) - V(s_t)|}{d(s_t, s_{t+1})},
\end{equation}
where $d$ is a function to measure the distance between two adjacent states. Here, we use the cosine similarity of representations from the last hidden state and the last token position. We sample a subset of 1000 trajectories from $\mathcal{D_{\text{PRM}}}$ and compare the constant calculated with state values from \model versus ScalarPRM. Empirically, a smaller number of $L_{\text{smoothness}}$ indicates a smoother PRM.

\section{Study of \model}
\subsection{Reward Smoonthness}
To illustrate the comparison of state values obtained from the \model and the ScalarPRM, we focus on the difference in their estimates across different quantile bins of $V_\text{Scalar}$.
The x-axis represents quantile bins of $V_\text{Scalar}$, which divides the range of state values computed by the Scalar PRM method into intervals. The y-axis depicts the average difference between the state values derived from TDRM and Scalar PRM, defined as Avg. ($V_\text{TDRM}$ - $V_\text{Scalar}$). A negative value on the y-axis indicates that TDRM estimates lower state values compared to ScalarPRM in the corresponding quantile bin, while values closer to zero indicate smaller differences between the two methods.

\begin{figure*}[th!]
    \centering
    \includegraphics[width=0.6\linewidth]{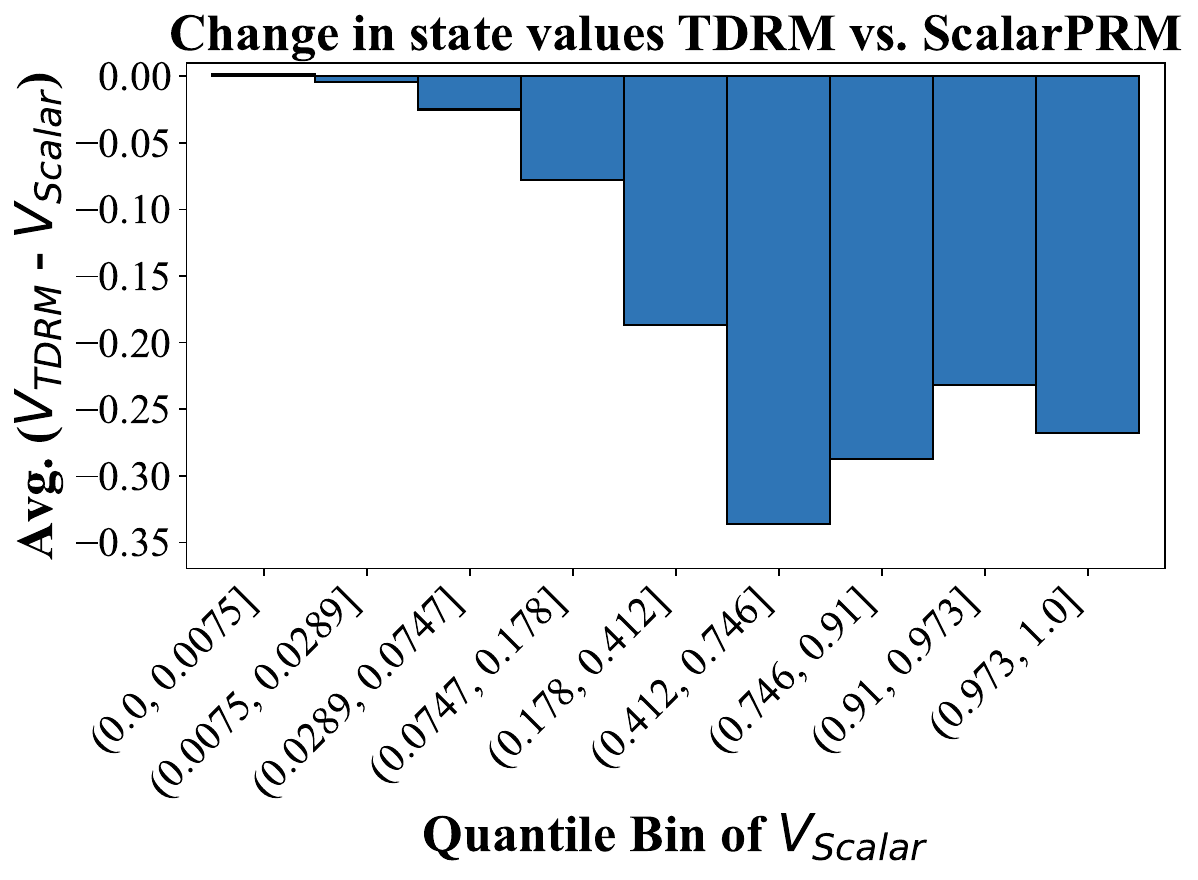}
    \caption{Comparison of state values from the \model and the ScalarPRM.}
    \label{fig: reward_smooth}
\end{figure*}

From the Figure~\ref{fig: reward_smooth}, it can be observed that:

$\bullet$ Lower Quantile Bins (e.g., (0.0, 0.0075)): The average state value difference is close to zero, meaning that TDRM and ScalarPRM compute nearly identical state values for smaller $V_\text{Scalar}$ values.

$\bullet$ Higher Quantile Bins (e.g., (0.412, 0.746) and beyond): The difference becomes significantly negative, indicating that \model tends to substantially reduce the state value for states that are assigned larger $V_\text{Scalar}$ values by ScalarPRM.

In conclusion, the ability of the \model to reduce the values of high-reward states can be instrumental in achieving smoother rewards during process reward model training or promoting policy robustness in reinforcement learning tasks.

\begin{table}[ht!]
    \centering
    \caption{Ablation study of 
    general models and reasoning models on the mathematical benchmarks.}
    \resizebox{0.92\textwidth}{!}{  
    \begin{tabular}{lllllllc}
        \specialrule{.16em}{0pt}{.65ex}  
        Model & \begin{tabular}[c]{@{}l@{}}Data \\ Size\end{tabular}                           & \begin{tabular}[c]{@{}l@{}}MATH \\ 500\end{tabular}                     & \begin{tabular}[c]{@{}l@{}}Minerva \\
        Math\end{tabular} & \begin{tabular}[c]{@{}l@{}}Olympiad \\ Bench\end{tabular} & \begin{tabular}[c]{@{}l@{}}AIME24 \\ (Pass@1)\end{tabular} & AMC23                        & Avg.       \\ 
        \specialrule{.10em}{.4ex}{.65ex}  
        \multicolumn{8}{c}{\textbf{Backbone is Base Model, Qwen Series}} \\
        \specialrule{.10em}{.4ex}{.65ex}  
        \multicolumn{1}{l}{\textbf{Qwen2.5-0.5B}}        & \multicolumn{1}{l}{-}         &15.8	&4.8	&2.8	&0.0	&12.5	&7.2         \\
        \specialrule{.10em}{.4ex}{.65ex}  
        + Rule-based    & 2.5k 	&29.8	&4.0	&7.0	&0.0	&12.5	&\underline{10.7} \\
        + w/ PRM & 2.5k &7.8	&1.5	&1.5	&0.0	&7.5 &3.7 \\
        \rowcolor{table-blue!66} + Ours             & 2.5k 	&26.2	&4.8	&7.1	&0.0	&15.0	&\textbf{10.8} (\textbf{\textcolor{darkred}{+0.9\%}}) \\
        \specialrule{.10em}{.4ex}{.65ex}
        \multicolumn{1}{l}{\textbf{Qwen2.5-1.5B}}        & \multicolumn{1}{l}{-}         &29.6	&6.6	&6.5	&0.0	&12.5	&11.0  \\ 
        \specialrule{.10em}{.4ex}{.65ex} 
        + Rule-based    & 2.5k &58.0	&12.1	&18.7	&0.0	&27.5 &\underline{23.3} \\ 
        + w/ PRM & 2.5k &16.0	&4.8	&5.9	&0.0	&15.0	&8.3 \\
        \rowcolor{table-blue!66} + Ours             & 2.5k 	&52.8	&9.9	&17.8	&3.3	&35.0 &\textbf{23.8} (\textbf{\textcolor{darkred}{+2.2\%}}) \\
        \specialrule{.10em}{.4ex}{.65ex}  
        \multicolumn{8}{c}{\textbf{Backbone is Chat Model, GLM Series}}\\
        \specialrule{.10em}{.4ex}{.65ex} 
        \textbf{GLM4-9B-0414} &- &65.8	&36.8	&28.7	&10.0	&42.5  &36.8 \\
        \specialrule{.10em}{.4ex}{.65ex} 
        + Rule-based &2.5k &74.0	&38.6	&36.3	&6.7	&47.5  &\underline{40.6} \\ 
        w/ PRM &2.5k &68.2	&39.3	&33.5	&10.0	&35.0		&37.2 \\
        \rowcolor{table-blue!66} + Ours    &2.5k &72.2	&37.1	&32.0	&20.0	&47.5	&\textbf{41.8} (\textbf{\textcolor{darkred}{+3.0\%}})  \\
        \specialrule{.10em}{.4ex}{.65ex}  
        \multicolumn{8}{c}{\textbf{Backbone is Reasoning Model, GLM Series}}\\
        \specialrule{.10em}{.4ex}{.65ex} 
        \textbf{GLM-Z1-9B-0414} &- &93.6	&43.8	&65.5	&73.3	&92.5 &73.7 \\
        \specialrule{.10em}{.4ex}{.65ex}  
        + Rule-based &2.5k &95.6	&43.4	&65.2	&73.3	&97.5 &75.0\\ 
        + w/ PRM &2.5k &95.6	&47.4	&67.0	&70.0	&97.5 &\underline{75.5} \\ 
        \rowcolor{table-blue!66} + Ours    &2.5k &94.6	&44.9	&66.5	&80.0	&97.5  &\textbf{76.7} (\textbf{\textcolor{darkred}{+1.6\%}}) \\
        \specialrule{.10em}{.4ex}{.65ex}  
        \multicolumn{8}{c}{\textbf{Backbone is Base Model, Qwen-Math Series}} \\
        \specialrule{.10em}{.4ex}{.65ex}
        \multicolumn{1}{l}{\textbf{Qwen2.5-Math-1.5B}}        & \multicolumn{1}{l}{-}         &42.2	&8.8	&27.0	&10.0	&37.5 &25.1  \\ 
        \specialrule{.10em}{.4ex}{.65ex} 
        + Rule-based    & 2.5k &67.6	&21.3	&31.0	&6.7	&52.5 &\underline{35.8} 
        \\ 
        +w/ PRM & 2.5k &63.8	&19.9	&26.7	&16.7	&50.0  &35.4 \\ 
        \rowcolor{table-blue!66} + Ours             & 2.5k 	&66.2	&18.4	&30.1	&13.3	&55.0 &\textbf{36.6} (\textbf{\textcolor{darkred}{+2.1\%}}) \\
        \specialrule{.10em}{.4ex}{.65ex}
        \multicolumn{1}{l}{\textbf{Qwen2.5-Math-7B}}        & \multicolumn{1}{l}{-}           & {63.6}  & {12.5}    & {25.8}      & {13.3}       & {42.5}  & 31.5         \\ 
        + Our Template & \multicolumn{1}{l}{-}         & {68.8}  & {16.2}    & {31.1}      & {13.3}       & {62.5}  & 38.4         \\ 
        \specialrule{.10em}{.4ex}{.65ex}  
        + Rule-based            & 2.5k & 73.2& 25.0               & 37.8               & 23.3                & 65.0  & \underline{44.9} \\
        + w/ PRM & 2.5k &58.0	&22.4	&20.6	&6.7	&30.0 &27.5 \\
        \rowcolor{table-blue!66} + Ours             & 2.5k & 74.6 & 26.8   & 37.3     & 36.7      & 62.5 & \textbf{47.6} (\textbf{\textcolor{darkred}{+6.0\%}}) \\
        \specialrule{.10em}{.4ex}{.65ex} 
        \multicolumn{8}{c}{\textbf{Backbone is Reasoning Model, DeepSeek Series}}\\
        \specialrule{.10em}{.4ex}{.65ex} 
        \textbf{DS-R1-Distill-Qwen-1.5B} &- &70.6	&26.5	&32.1	&16.7	&50.0  &39.2 \\
        \specialrule{.10em}{.4ex}{.65ex} 
        + Rule-based &2.5k &75.4	&26.8	&36.1	&20.0	&57.5 &\underline{43.2} \\
        + w/ PRM &2.5k &69.8 &18.0 &30.5 &33.3 &45.0 &39.3\\
        \rowcolor{table-blue!66} + Ours &2.5k &79.8	&30.5	&38.2	&30.0	&70.0  &\textbf{49.7} (\textbf{\textcolor{darkred}{+15.0\%}}) \\
        \specialrule{.10em}{.4ex}{.65ex} 
        \textbf{DS-R1-Distill-Qwen-7B} &- &88.0	&43.0	&49.9	&63.3	&82.5  &\underline{65.3} \\
        \specialrule{.10em}{.4ex}{.65ex} 
        + Rule-based &2.5k &89.6	&46.0	&52.4	&50.0	&82.5  &64.1 \\
        + w/ PRM &2.5k &84.2 &44.5 &45.8 &46.7 &77.5 &59.7 \\
        \rowcolor{table-blue!66} + Ours    &2.5k &91.8	&50.4	&54.1	&53.3	&87.5  &\textbf{67.4} (\textbf{\textcolor{darkred}{+3.2\%}}) \\
         \specialrule{.16em}{.4ex}{0pt}
    \end{tabular}
    }
    \label{tab: ablation_study}
\end{table}
\subsection{Ablation Study of Training Models}
\vpara{Accuracy and Reward Score in RL.}
To further demonstrate the effectiveness of \model in RL training, we show a comparison between \model and checkpoints trained with pure verifiable reward. We compare the verifiable reward during training for \model and the pure rule-based method in Table~\ref{tab: ablation_study}.
We can observe that:

$\bullet$ Effectiveness of Rule-based Approaches. Adding rule-based methods consistently improves average performance. For example, 
% in the ``Qwen2.5-0.5B'' model, the average accuracy rises from 7.2\% to 10.7\% (+3.5\%), and 
in ``DS-R1-Distill-Qwen-1.5B'', the accuracy increases from 39.2\% to 43.2\% (+4.0\%). This demonstrates the critical role of the rule-based method in mathematical reasoning.

$\bullet$ Impact of PRM trained with TD. The addition of our trained PRM further enhances performance over the purely rule-based approach, especially for larger models. For instance, in ``Qwen2.5-Math-1.5B'', the average improves from 35.8\% (Rule-based) to 36.6\% (+0.8\%) with trained PRM. This shows PRM's ability to refine value estimation and align intermediate reasoning steps.

$\bullet$ Performance of Ours. Combining rule-based methods and PRM (``Ours'') achieves the best results across all settings. Notably, ``DS-R1-Distill-Qwen-7B'' reaches an average accuracy of 67.4\% with a relative improvement of 3.2\%, and ``Qwen2.5-Math-1.5B'' improves by +2.1\%, highlighting the synergy of the two components in enhancing smoothness and temporal consistency.

\section{Case Study}
\begin{tcolorbox}[colframe=custom_blue, title=An Example of Sampled Answer with Reward During RL Training, colback=white]
\footnotesize{
    \textbf{Question:}  Evaluate $\lfloor{\sqrt{12}}\rfloor^2$\\

    \textbf{Answer from LLM:}\\

    \textit{\textcolor{blue}{→ Clear plan provided before calculation.}}\\
    To solve the problem, we need to break it down into a sequence of steps:\\\\
    1. Compute \(\sqrt{12}\).\\
    2. Compute the floor value of \(\sqrt{12}\), denoted as \(\lfloor \sqrt{12} \rfloor\).\\
    3. Square the result from step 2.\\

    \textit{\textcolor{blue}{→ Code is used to verify and trace all intermediate values.}}\\
    Let's go through each step using Python code to ensure accuracy.\\

    \begin{tcolorbox}[colback=white, colframe=custom_blue, title=Python Code]
    \textit{\textcolor{blue}{→ Clean and well-commented code for reproducibility.}}\\
    \texttt{import math}\\
    \texttt{\# Step 1: Compute the square root of 12}\\
    \texttt{sqrt\_12 = math.sqrt(12)}\\

    \texttt{\# Step 2: Compute the floor value of the square root of 12}\\
    \texttt{floor\_sqrt\_12 = math.floor(sqrt\_12)}\\

    \texttt{\# Step 3: Square the floor value}\\
    \texttt{result = floor\_sqrt\_12 ** 2}\\

    \texttt{print((sqrt\_12, floor\_sqrt\_12, result))}
    \end{tcolorbox}

    \textit{\textcolor{blue}{→ Output includes intermediate values, adding transparency.}}\\
    \texttt{(3.4641016151377544, 3, 9)}\\

    \textit{\textcolor{blue}{→ Step-by-step interpretation of output builds trust in the logic.}}\\
    Here is the step-by-step result:\\\\
    1. \(\sqrt{12} \approx 3.464\).\\
    2. \(\lfloor \sqrt{12} \rfloor = 3\).\\
    3. \(3^2 = 9\).\\

    \textit{\textcolor{blue}{→ Final boxed result clearly highlights the answer.}}\\
    Thus, the final answer is: \(\boxed{9}\)\\

    \textbf{Reward:} 2.58
    \textit{\textcolor{blue}{→ High reward reflects well-structured and interpretable reasoning.}}
}
\end{tcolorbox}
\normalsize

\end{document}